\documentclass[final,3p,times,twocolumn]{elsarticle}

\usepackage{hyperref}
\usepackage{amssymb}
\usepackage{bm}
\usepackage{amsmath}
\usepackage{graphicx}
\usepackage{subfigure}
\usepackage{booktabs}
\usepackage{multirow}
\usepackage{url}
\makeatletter
\g@addto@macro{\UrlBreaks}{\UrlOrds}
\makeatother
\usepackage{algorithm}
\usepackage{algorithmic}
\usepackage{color}
\biboptions{numbers,sort&compress}
\usepackage{lineno}

\journal{Knowledge-Based Systems}
\newcommand{\diff}{}
\newcommand{\revise}{}

\begin{document}

\begin{frontmatter}



\title{adVAE: a Self-adversarial Variational Autoencoder \\with Gaussian Anomaly Prior Knowledge \\for Anomaly Detection\tnoteref{conflicts}}
\tnotetext[conflicts]{No author associated with this paper has disclosed any potential or pertinent conflicts which may be perceived to have impending conflict with this work.}


\author[SJTU]{Xuhong Wang\corref{cor1}}
\ead{wang\_xuhong@sjtu.edu.cn}
\author[SJTU]{Ying Du}
\ead{duying@sjtu.edu.cn}
\author[WHU]{Shijie Lin}
\ead{linshijie@whu.edu.cn}
\author[SJTU]{Ping Cui}
\ead{cuiping@sjtu.edu.cn}
\author[UCD]{Yuntian Shen}
\ead{ytshen@ucdavis.edu}
\author[SJTU]{Yupu Yang}
\ead{ypyang@sjtu.edu.cn}
\cortext[cor1]{Corresponding author}
\address[SJTU]{Shanghai Jiao Tong University, Shanghai, China}
\address[WHU]{Wuhan University, Wuhan, China}
\address[UCD]{University of California, Davis , U.S.A.}

\begin{abstract}
Recently, deep generative models have become increasingly popular in unsupervised anomaly detection. However, deep generative models aim at recovering the data distribution rather than detecting anomalies. Moreover, deep generative models have the risk of overfitting training samples, which has disastrous effects on anomaly detection performance. To solve the above two problems, we propose a self-adversarial variational autoencoder (adVAE) with a Gaussian anomaly prior assumption. We assume that both the anomalous and the normal prior distribution are Gaussian and have overlaps in the latent space. Therefore, a Gaussian transformer net T is trained to synthesize anomalous but near-normal latent variables. Keeping the original training objective of a variational autoencoder, a generator G tries to distinguish between the normal latent variables encoded by E and the anomalous latent variables synthesized by T, and the encoder E is trained to discriminate whether the output of G is real. These new objectives we added not only give both G and E the ability to discriminate, but also become an additional regularization mechanism to prevent overfitting. Compared with other competitive methods, the proposed model achieves significant improvements in extensive experiments. The employed datasets and our model are available in a Github repository.
\end{abstract}


\end{frontmatter}


\section{Introduction}
Anomaly detection (or outlier detection) can be regarded as the task of identifying rare data items that differ from the majority of the data. Anomaly detection is applicable in a variety of domains, such as intrusion detection, fraud detection, fault detection, health monitoring, and security checking~\cite{osada2017network,DBLP:journals/jnca/AbdallahMZ16,cui2019improved,schlegl2017unsupervised,DBLP:conf/accv/AkcayAB18}. \revise{Owing to the lack of labeled anomaly samples, there is a large skew between normal and anomaly class distributions. Some attempts~\cite{DBLP:journals/kbs/ZhangBXRFQF19,zhou2019deep,JMLR:v18:16-365} use several imbalanced-learning methods to improve the performance of supervised anomaly detection models.} Moreover, unsupervised models are more popular than supervised models in the anomaly detection field. Reference \cite{pimentel2014review} reviewed machine-learning-based anomaly detection algorithms comprehensively. 

Recently, deep generative models have become increasingly popular in anomaly detection~\cite{chalapathy2019deep}. \diff{A generative model can learn a probability distribution model by being trained on an anomaly-free dataset. Afterwards, outliers can be detected by their deviation from the probability model. The most famous deep generative models are variational autoencoders (VAEs)~\cite{kingma2013auto} and generative adversarial networks (GANs)~\cite{goodfellow2014generative}. }

VAEs have been used in many anomaly detection studies. The first work using VAEs for anomaly detection~\cite{an2015variational} declared that VAEs generalize more easily than autoencoders (AEs), because VAEs work on probabilities. \cite{8279425,suh2016echo} used different types of RNN-VAE architectures to recognize the outliers of time series data. \cite{osada2017network} and \cite{xu2018unsupervised} implemented VAEs for intrusion detection and internet server monitoring, respectively. Furthermore, GANs and adversarial autoencoders~\cite{makhzani2015adversarial} have also been introduced into image~\cite{schlegl2017unsupervised,pidhorskyi2018generative} and video~\cite{ravanbakhsh2017abnormal} anomaly detection. 

However, there are two serious problems in deep-generative-model-based anomaly detection methods.

(1) Deep generative models only aim at recovering the data distribution of the training set, which has an indirect contribution to detecting anomalies. Those earlier studies paid little attention to customize their models for anomaly detection tasks. Consequently, there is an enormous problem that those models only learn from available normal samples, without attempting to discriminate the anomalous data. Owing to the lack of discrimination, it is hard for such models to learn useful deep representations for anomaly detection tasks. 

(2) \diff{Plain VAEs use the regularization of the Kullback--Leibler divergence (KLD) to limit the capacity of the encoder, but there is no regularization implemented in the generator. Because neural networks are universal function approximators, generators can, in theory, cover any probability distribution, even without dependence on the latent variables \cite{DBLP:conf/iclr/0022KSDDSSA17}. However, previous attempts \cite{DBLP:conf/nips/FraccaroSPW16,DBLP:conf/aaai/SerbanSLCPCB17} have found it hard to benefit from using an expressive and powerful generator. \cite{DBLP:conf/iclr/0022KSDDSSA17} uses Bits-Back Coding theory \cite{honkela2004variational} to explain this phenomenon: if the generator can model the data distribution $p_{data}(x)$ without using information from the latent code $z$, it is more inclined not to use $z$. In this case, to reduce the KLD cost, the encoder loses much information about $x$ and maps $p_{data}(x)$ to the simple prior $p(z)$ (e.g., $\mathcal { N }(0,I)$) rather than the true posterior $p(z|x)$. Once this undesirable training phenomenon occurs, VAEs tend to overfit the distribution of existing normal data, which leads to a bad result (e.g., a high false positive rate), especially when the data are sparse. Therefore, the capacity of VAEs should be limited by an additional regularization in anomaly detection tasks.} 
\begin{figure}[htb!]
	\centering
	\subfigure[]
	{ 
		\label{fig:ouranomalyprior}
		\includegraphics[width=0.4\linewidth]{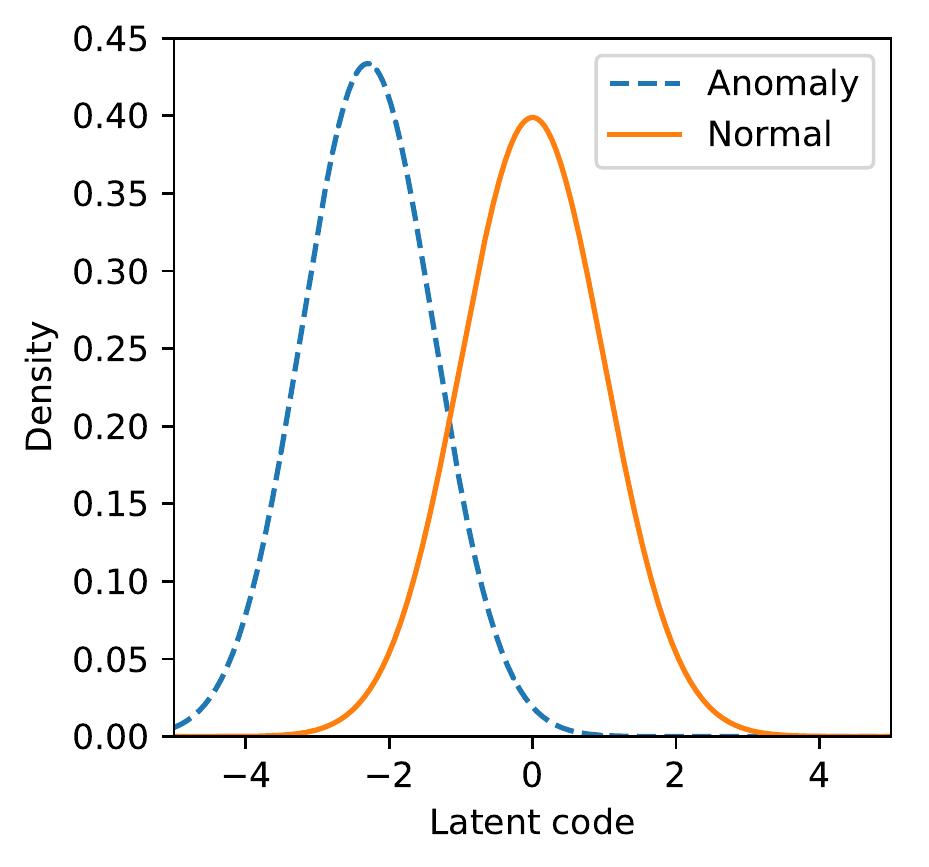}
	}
	\subfigure[]
	{
		\label{fig:self-adversarial}
		\includegraphics[width=0.5\linewidth]{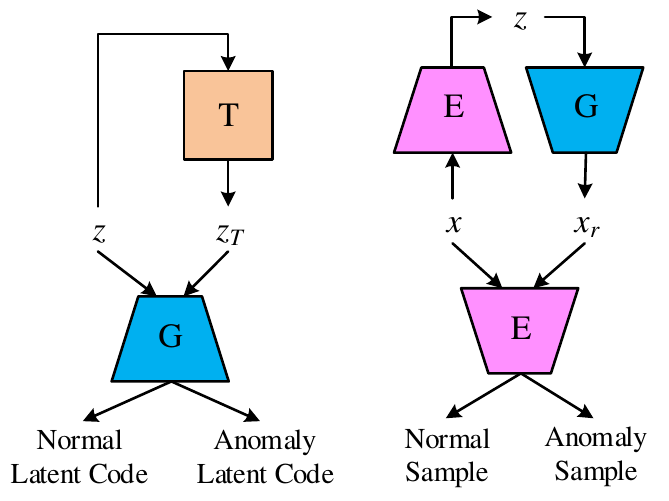}
	}
	\caption{(a) Our assumption is that the normal data prior distribution is a Gaussian distribution close to $\mathcal { N }(0,I)$, and the anomalous prior is another Gaussian distribution, whose mean and variance are unknown and different from the normal prior, with overlaps in the latent space. \diff{(b) This figure illustrates the basic structure of our self-adversarial mechanism. We propose additional discrimination objectives for both the encoder and the generator by adding two competitive relationships. $x$ and $z$ are normal items because of the anomaly-free training dataset. $x_r$ and $z_{T}$ can be regarded as anomalous item. $G$ is trained to distinguish $z$ and $z_{T}$, and $E$ tries to discern $x$ and $x_r$.}
	}
	\label{fig:introduction}
\end{figure}

There are only a handful of studies that attempt to solve the above two problems. 
\diff{MO-GAAL method~\cite{8668550} made the generator of a GAN stop updating before convergence in the training process and used the nonconverged generator to synthesize outliers. Hence, the discriminator can be trained to recognize outliers in a supervised manner. To try to solve the mode collapse issue of GANs, \cite{8668550} expands the network structure from a single generator to multiple generators with different objectives.}
\cite{kawachi2018complementary} proposed an assumption that the anomaly prior distribution is the complementary set of the normal prior distribution in latent space. However, \diff{we believe that this assumption may not hold true.} If the anomalous and the normal data have complementary distributions, which means that they are separated in the latent space, then we can use a simple method (such as KNN) to detect anomalies and achieve satisfactory results, but this is not the case. Both normal data and outliers are generated by some natural pattern. Natural data ought to conform to a common data distribution, and it is hard to imagine a natural pattern that produces such a strange distribution. 

To enhance deep generative models to distinguish between normal and anomalous samples and to prevent them from overfitting the given normal data, we propose a self-\textbf{ad}versarial \textbf{V}ariational \textbf{A}uto\textbf{e}ncoder (adVAE) with a Gaussian anomaly prior assumption and a self-adversarial regularization mechanism. \diff{The basic idea of this self-adversarial mechanism is adding discrimination training objectives to the encoder and the generator through adversarial training. These additional objectives will solve the above two problems at the same time; the details are as follows.} 

\diff{The encoder can be trained to discriminate the original sample and its reconstruction, but we do not have any anomaly latent code to train the generator. To synthesize the anomaly latent code, we propose a Gaussian anomaly hypothesis to describe the relationship between normal and anomaly latent space.} Our assumption is described in Figure~\ref{fig:ouranomalyprior}; both the anomalous and the normal prior distributions are Gaussian and have overlaps in the latent space. It is an extraordinarily weak and reasonable hypothesis, because the Gaussian distribution is the most widespread in nature.
\diff{The basic structure of our self-adversarial mechanism is shown in Figure~\ref{fig:self-adversarial}. The encoder is trained to discriminate the original sample $x$ and its reconstruction $x_r$,} and the generator tries to distinguish between the normal latent variables $z$ encoded by the encoder and the anomalous ones $z_{T}$ synthesized by $T$. These new objectives we added not only give $G$ and $E$ the ability to discern, but also introduce an additional regularization to prevent model overfitting. 

Our training process can be divided into two steps: (1) $T$ tries to mislead the generator $G$; meanwhile $G$ works as a discriminator. (2) $G$ generates realistic-like samples, and the encoder $E$ acts as a discriminator of $G$. To make the training phase more robust, inspired by~\cite{huang2018introvae}, we train alternatively between the above two steps in a mini-batch iteration.

Our main contributions are summarized as follows:
\begin{enumerate}
    \item We propose a novel and important concept that deep generative models should be customized to learn to discriminate outliers rather than being used in anomaly detection directly without any suitable customization.
    \item We propose a novel self-adversarial mechanism, which is the prospective customization of a plain VAE, enabling both the encoder and the generator to discriminate outliers.
    \item The proposed self-adversarial mechanism also provides a plain VAE with a novel regularization, which can significantly help VAEs to avoid overfitting normal data.
    \item We propose a Gaussian anomaly prior knowledge assumption that describes the data distribution of anomalous latent variables. Moreover, we propose a Gaussian transformer net $T$ to integrate this prior knowledge into deep generative models. 
    \item The decision thresholds are automatically learned from normal data by a kernel density estimation (KDE) technique, whereas earlier anomaly detection works (e.g., \cite{an2015variational,xu2018unsupervised}) often omit the importance of automatic learning thresholds.
\end{enumerate}

\section{Preliminary}
\subsection{Conventional Anomaly Detection}
Anomaly detection methods can be broadly categorized into probabilistic, distance-based, boundary-based and reconstruction-based.

(1) Probabilistic approach, such as GMM \cite{DBLP:conf/icpr/IlonenPKK06} and KDE \cite{DBLP:conf/icpr/YeungC02}, uses statistical methods to estimate the probability density function of the normal class. A data point is defined as an anomaly if it has low probability density. (2) Distance-based approach has the assumption that normal data are tightly clustered, while anomaly data occur far from their nearest neighbours. These methods depend on the well-defined similarity measure between two data points. The basic distance-based methods are LOF \cite{DBLP:conf/sigmod/BreunigKNS00} and its modification \cite{DBLP:journals/ijon/TangH17}. (3) Boundary-based approach, mainly involving OCSVM~\cite{OCSVM} and SVDD~\cite{DBLP:journals/kbs/YinWF18}, typically try to define a boundary around the normal class data. Whether the unknown data is an anomaly instance is determined by their location with respect to the boundary. (4) Reconstruction-based approach assumes that anomalies are incompressible and thus cannot be effectively reconstructed from low-dimensional projections. In this category, PCA \cite{Olive2017} and its variations \cite{DBLP:journals/candie/HarrouKCTS15,baklouti2016iterated} are widely used, effective techniques to detect anomalies. \diff{Besides, AE and VAE based methods also belong to this category, which will be explained detailedly in the next two subsections.}
\subsection{Autoencoder-based Anomaly Detection}
An AE, which is composed of an encoder and a decoder, is a neural network used to learn reconstructions as close as possible to its original inputs. Given a datapoint $x\in \bm{R}^d$ ($d$ is the dimension of $x$), the loss function can be viewed as minimizing the reconstruction error between the training data and the outputs of the AE, and $\theta$ and $\phi$ denote the hidden parameters of the encoder $E$ and the decoder $G$:
\begin{linenomath}
\begin{equation}
\mathcal { L } _ { AE } ( x , \phi , \theta ) = \left\| x - G _ { \theta } \left( E _ { \phi } ( x ) \right) \right\|  ^ { 2 }
\end{equation}
\end{linenomath}

After training, the reconstruction error of each test data will be regarded as the anomaly score. The data with a high anomaly score will be defined as anomalies, because only the normal data are used to train the AE. The AE will reconstruct normal data very well, while failing to do so with anomalous data that the AE has not encountered.
\subsection{VAE-based Anomaly Detection}
\label{VAE}
The net architecture of VAEs is similar to that of AEs, with the difference that the encoder of VAEs forces the representation code $z$ to obey some type of prior probability distribution ${p}({z})$ (e.g., $\mathcal { N }(0,I)$). Then, the decoder generates new realistic data with code $z$ sampled from ${p}({z})$. In VAEs, both the encoder and decoder conditional distributions are denoted as $q _ { \phi } (  { z } |  { x } )$ and $p _ { \theta } (  { z } |  { x } )$. The data distribution $p _ { \theta } ( {x} )$ is intractable by analytic methods, and thus variational inference methods are introduced to solve the maximum likelihood $\log p _ { \theta } (  { x } )$:
\begin{linenomath}
\begin{equation}
\begin{aligned} 
    \mathcal { L } (  { x } ) & \!=\!  \log p _ { \theta } (  { x } )  -  { KL } \left[ q _ { \phi } (  { z } |  { x } ) \| p _ { \theta } (  { z } |  { x } ) \right]  \\
     & \!=\!  { E } _ { q _ { \phi } (  { z } |  { x } ) } \left[ \log p _ { \theta } (  { x } ) \!+\! \log p _ { \theta } (  { z } |  { x } ) \!-\! \log q _ { \phi } (  { z } |  { x } ) \right] \\ 
     & \!=\!- KL  \left( q _ { \phi } ( z | x ) \| p _ { \theta } ( z ) \right) + E _ { q _ { \phi } ( z | x ) } \left[ \log p _ { \theta } ( x | z ) \right]
\end{aligned}
\label{equation:ELBO}
\end{equation}
\end{linenomath}

KLD is a similarity measure between two distributions. To estimate this maximum likelihood, VAE needs to maximize the evidence variational lower bound (ELBO) $\mathcal { L } (  { x } )$.
To optimize the KLD between $ q _ { \phi }( z | x )$ and $ p _ { \theta }  (z )$, the encoder estimates the parameter vectors of the Gaussian distribution $ q _ { \phi }( z | x )$ : mean~$\mu$ and standard deviation~$\sigma$. There is an analytical expression for their KLD, because both $ q _ { \phi }( z | x )$ and $ p _ { \theta }  (z )$ are Gaussian. To optimize the second term of equation (\ref{equation:ELBO}), VAEs minimize the reconstruction errors between the inputs and the outputs. Given a datapoint $x\in \bm{R}^d$, the objective function can be rewritten as
\begin{linenomath}
\begin{equation}
\begin{aligned} 
\mathcal { L } _ { VAE }=&\mathcal { L } _ { MSE } (x,G_ { \theta }( z ))+\lambda\mathcal { L } _ { KLD }\left(E_ { \phi } \left(x\right)\right)\\=&\mathcal { L } _ { MSE } (x,x_{r})+\lambda\mathcal { L } _ { KLD }(\mu,\sigma)
\end{aligned}
\label{equation:VAELOSS}
\end{equation}
\begin{equation}
\begin{aligned} 
\mathcal { L } _ { MSE }(x,x_{r})=&\left\| x - x_{r} \right\| ^ { 2 } 
\end{aligned}
\label{equation:MSELOSS}
\end{equation}
\begin{equation}
\begin{aligned} 
\mathcal { L } _ { KLD }(\mu,\sigma)=&KL  \left( q _ { \phi } ( z | x ) \| p _ { \theta } ( z ) \right)\\= & KL  \left( \mathcal { N } \left( z;  \mu ,  \sigma  ^ { 2 } \right) \| \mathcal { N } \left( z; 0  ,  I  \right) \right)\\=&\int \mathcal { N } \left(  z; \mu  ,  \sigma ^ { 2 }  \right)  \log\frac{\mathcal { N } \left( z;  \mu  ,  \sigma ^ { 2 }  \right)}{\mathcal { N } ( z; 0,I)} dz \\=&\frac { 1 } { 2 }  \left( 1+ \log   (\sigma   ^ { 2 })  -  \mu  ^ { 2 } -  \sigma ^ { 2 } \right)
\end{aligned}
\label{equation:KLDLOSS}
\end{equation}
\end{linenomath}

The first term $\mathcal { L } _ { MSE }(x,x_{r})$ is the mean squared error (MSE) between the inputs and their reconstructions. The second term $\mathcal { L } _ { KLD }(\mu,\sigma)$ regularizes the encoder by encouraging the approximate posterior $ q _ { \phi }( z | x )$ to match the prior $ p (z )$. \diff{To hold the tradeoff between these two targets, each KLD target term is multiplied by a scaling hyperparameter $\lambda$.}

AEs define the reconstruction error as the anomaly score in the test phase, whereas 
VAEs use the reconstruction probability~\cite{an2015variational} to detect outliers. To estimate the probabilistic anomaly score, VAEs sample $z$ according to the prior $ p _ { \theta }  (z )$ for $L$ times and calculate the average reconstruction error as the reconstruction probability. This is why VAEs work more robustly than traditional AEs in the anomaly detection domain.
\subsection{GAN-based Anomaly Detection}
Since GANs~\cite{goodfellow2014generative} were first proposed in 2014, GANs have become increasingly popular and have been applied for diverse tasks. A GAN model comprises two components, which contest with each other in a cat-and-mouse game, called the generator and discriminator. The generator creates samples that resemble the real data, while the discriminator tries to recognize the fake samples from the real ones. \diff{The generator of a GAN synthesizes informative potential outliers to assist the discriminator in describing a boundary that can separate outliers from normal data effectively~\cite{8668550}.} 
When a sample is input into a trained discriminator, the output of the discriminator is defined as the anomaly score. \diff{However, suffering from the mode collapsing problem, GANs usually could not learn the boundaries of normal data well, which reduces the effectiveness of GANs in anomaly detection applications. To solve this problem, \cite{8668550} propose MOGAAL and suggest stopping optimizing the generator before convergence and expanding the network structure from a single generator to multiple generators with different objectives. \revise{In addition, WGAN-GP~\cite{DBLP:conf/nips/GulrajaniAADC17}, one of the most advanced GAN frameworks, proposes a Wasserstein distance and gradient penalty trick to avoid mode collapsing. In our experiments, we also compared the anomaly detection performance between a plain GAN and WGAN-GP.}}

\section{Self-adversarial Variational Autoencoder}
\begin{figure*}[htb!]
    \centering
    \includegraphics[width=0.8\linewidth]{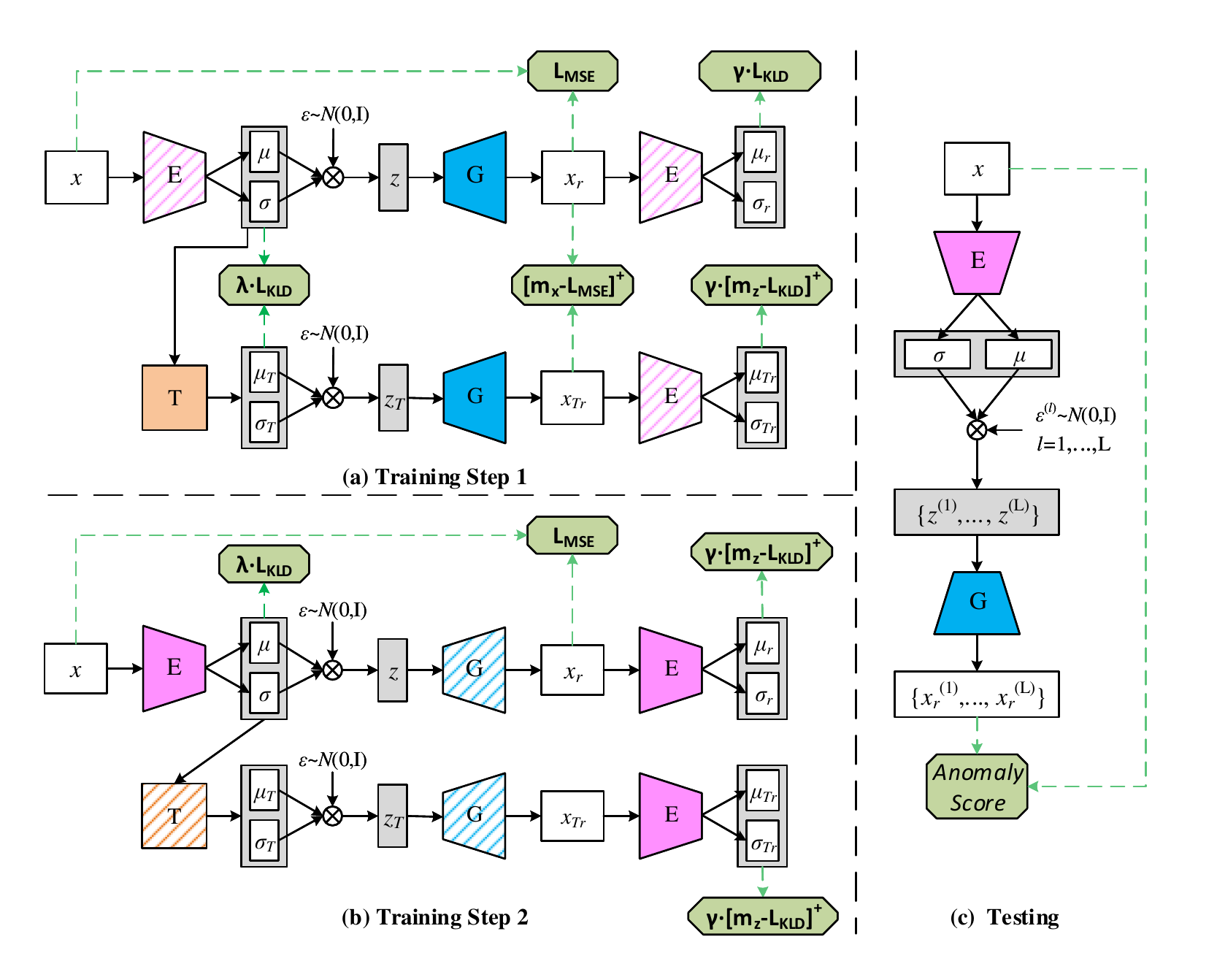}
    \caption{Architecture and training flow of adVAE. \diff{As adVAE is a variation of plain VAE, $\lambda$ is the hyperparameter derived from VAE. We add discrimination objectives to both the encoder and the generator by adversarial learning. The larger the $\gamma$ or $m_z$, the larger the proportion of the encoder discrimination objective in the total loss. The larger the $m_x$, the larger the proportion of the generator discrimination objective.} (a) Updating the decoder $G$ and the Gaussian transformer $T$. (b) Updating the encoder $E$. (c) Anomaly score calculation. Hatch lines indicate that the weights of the corresponding networks are frozen.}
    \label{fig:adVAE}
\end{figure*}
In this section, a self-\textbf{ad}versarial \textbf{V}ariational \textbf{A}uto\textbf{e}ncoder (adVAE) for anomaly detection is proposed. To customize plain VAE to fit anomaly detection tasks, we propose the assumption of a Gaussian anomaly prior and introduce the self-adversarial mechanism into traditional VAE. The proposed method consists of three modules: an encoder net ${E}$, a generative net ${G}$, and a Gaussian transformer net ${T}$.

There are two competitive relationships in the training phase of our method: (1) To generate a potential anomalous prior distribution and enhance the generator's ability to discriminate between normal and anomalous priors, we train the Gaussian transformer ${T}$ and the generator ${G}$ with adversarial objectives simultaneously. 
(2) To produce more realistic samples in a competitive manner and make the encoder learn to discern, we train the generator and the encoder analogously to the generator and discriminator in GANs. 

According to equation~(\ref{equation:VAELOSS}), there are two components in the objective function of VAEs: $\mathcal { L } _ { MSE }$ and $\mathcal { L } _ { KLD }$. The cost function of adVAE is a modified combination objective of these two terms. In the following, we describe the training phase in subsections \ref{TandG} to \ref{Alternating} and subsections \ref{score} to \ref{detect} address the testing phase.
\subsection{Training Step 1: Competition between T and G}
\label{TandG}
The generator of plain VAE is often so powerful that it maps all the Gaussian latent code to the high-dimensional data space, even if the latent code is encoded from anomalous samples. Through the competition between ${T}$ and ${G}$, we introduce an effective regularization into the generator. 

Our anomalous prior assumption suggests that it is difficult for the generator of plain VAE to distinguish the normal and the anomalous latent code, because they have overlaps in the latent space. To solve this problem, we synthesize anomalous latent variables and make the generator discriminate the anomalous from the normal latent code. As shown in Figure \ref{fig:adVAE} (a), we freeze the weights of ${E}$ and update ${G}$ and ${T}$ in this training step. The Gaussian transformer ${T}$ receives the normal Gaussian latent variables $z$ encoded from the normal training samples as the inputs and transforms $z$ to the anomalous Gaussian latent variables $z_{T}$ with different mean $\mu_{T}$ and standard deviation $\sigma_{T}$. ${T}$ aims at reducing the KLD between $\{z;\mu,\sigma\}$ and $\{z_{T};\mu_{T},\sigma_{T}\}$, and ${G}$ tries to generate as different as possible samples from such two similar latent codes.

Given a datapoint $x\in \bm{R}^d$, the objective function in this competition process can be defined as
\begin{linenomath}
\begin{equation}
\begin{aligned} 
    \mathcal { L } _ { G } =\mathcal { L } _ { Gz } +\mathcal { L } _ {  Gz_{T}  }
\end{aligned}
\label{equation:GLOSS}  
\end{equation}
\begin{equation}
\begin{aligned} 
\mathcal { L } _ { Gz }=&\mathcal { L } _ { MSE }\left(x,G\left(z\right)\right) +\gamma\mathcal { L } _ { KLD }\left(E\left(G\left(z\right)\right)\right)\\=&\mathcal { L } _ { MSE }\left(x,x_{r}\right) +\gamma\mathcal { L } _ { KLD }(\mu_{r},\sigma_{r})
\end{aligned}
\label{equation:GZLOSS}  
\end{equation}
\begin{equation}
\begin{aligned} 
    \mathcal { L } _ { Gz_{T} } =&\left[ m_{x}-\mathcal { L } _ { MSE }(G\left(z\right),G\left(z_{T}\right))\right]^{+} \\ &+\gamma\left[ m_{z}-\mathcal { L } _ { KLD }\left(E\left(G\left(z_{T}\right)\right)\right)\right]^{+}\\ =&\left[ m_{x}-\mathcal { L } _ { MSE }(x_{r},x_{Tr})\right]^{+} \\ &+\gamma\left[ m_{z}-\mathcal { L } _ { KLD }(\mu_{Tr},\sigma_{Tr})\right]^{+}
\end{aligned}
\label{equation:GZTLOSS}  
\end{equation}
\begin{equation}
    \begin{aligned} 
        \mathcal { L } _ { T } &= KL  \left( \mathcal { N } \left(  z; \mu  ,  \sigma ^ { 2 }  \right)\|\mathcal { N } ( z; \mu_{T}  ,  \sigma_{T}^ { 2 }   ) \right)\\ 
        &= \int \mathcal { N } \left(  z; \mu  ,  \sigma ^ { 2 }  \right)  \log\frac{\mathcal { N } \left( z;  \mu  ,  \sigma ^ { 2 }  \right)}{\mathcal { N } ( z; \mu_{T}  ,  \sigma_{T}^ { 2 }   )} dz    \\ 
        &=\log \frac { \sigma _ { T } } { \sigma  } + \frac { \sigma  ^ { 2 } + \left( \mu  - \mu _ { T } \right) ^ { 2 } } { 2 \sigma _ { T } ^ { 2 } } - \frac { 1 } { 2 } 
    \end{aligned}
    \label{equation:TLOSS}  
    \end{equation}
\end{linenomath}

$[ \cdot ] ^ { + } = \max ( 0 , \cdot )$, $m_{x}$ is a positive margin of the MSE target, and $m_{z}$ is a positive margin of the KLD target. The aim is to hold the corresponding target term below the margin value for most of the time. $\mathcal { L } _ { Gz }$ is the objective for the data flow of $z$, and $\mathcal { L } _ { Gz_{T} }$ is the objective for the pipeline of $z_{T}$. 
  
$\mathcal { L } _ { T } $ and $\mathcal { L } _ { G } $ are two adversarial objectives, and the total objective function in this training step is the sum of the two: $\mathcal { L } _ { T } +\lambda\mathcal { L } _ { G } $. Objective $\mathcal { L } _ { T }$ encourages $T$ to mislead $G$ by synthesizing $z_{T}$ similar to $z$, such that $G$ cannot distinguish them. Objective $\mathcal { L } _ { G } $ forces the generator to distinguish between $z$ and $z_{T}$. $T$ hopes that $z_{T}$ is close to $z$, whereas $G$ hopes that $z_{T}$ is farther away from $z$. After iterative learning, $T$ and $G$ will reach a balance. $T$ will generate anomalous latent variables close to the normal latent variables, and the generator will distinguish them by different reconstruction errors. Although the anomalous latent variables synthesized by $T$ are not necessarily real, it is helpful for the models as long as they try to identify the outliers. 

Because the updating of $E$ will affect the balance of $T$ and $G$, we freeze the weights of $E$ when training $T$ and $G$. If we do not do this, it will be an objective of three networks' equilibrium, which is extremely difficult to optimize.
\subsection{Training Step 2: Training E like a Discriminator}
In the first training step demonstrated in the previous subsection, we freeze the weights of $E$. 
Instead, as shown in Figure~\ref{fig:adVAE} (b), we now freeze the weights of $T$ and $G$ and update the encoder $E$. 
The encoder not only attempts to project the data samples $x$ to Gaussian latent variables $z$ like the original VAE, but also works like a discriminator in GANs. The objective of the encoder is as follows:

\begin{equation}
\begin{aligned} 
    \mathcal { L } _ { E }=&\mathcal { L } _ { MSE }\left(x,G\left(z\right)\right)+\lambda\mathcal { L } _ { KLD }\left(E\left(x\right)\right)\\ &+\gamma\left[ m_{z}-\mathcal { L } _ { KLD }\left(E\left(G\left(z\right)\right)\right)\right]^{+}\\ &+\gamma\left[ m_{z}-\mathcal { L } _ { KLD }\left(E\left(G\left(z_{T}\right)\right)\right)\right]^{+}\\ =&\mathcal { L } _ { MSE }(x,x_{r})+\lambda\mathcal { L } _ { KLD }(\mu,\sigma)\\ &+\gamma\left[ m_{z}-\mathcal { L } _ { KLD }(\mu_{r},\sigma_{r})\right]^{+}\\ &+\gamma\left[ m_{z}-\mathcal { L } _ { KLD }(\mu_{Tr},\sigma_{Tr})\right]^{+}
\end{aligned}
\label{equation:ELOSS}  
\end{equation}

The first two terms of Equation~\ref{equation:ELOSS} are the objective function of plain VAE. The encoder is trained to encode the inputs as close to the prior distribution when the inputs are from the training dataset. The last two terms are the discriminating loss we proposed. The encoder is prevented from mapping the reconstructions of training data to the latent code of the prior distribution. 

The objective $\mathcal { L } _ { E }$ provides the encoder with the ability to discriminate whether the input is normal because the encoder is encouraged to discover differences between the training samples (normal) and their reconstructions (anomalous). It is worth mentioning that the encoder with discriminating ability also helps the generator distinguish between the normal and the anomalous latent code. 

\subsection{Alternating between the Above Two Steps}
\label{Alternating}
\begin{algorithm}[tb!]
  \caption{Training adVAE model}
  \label{alg:adVAE}
  
  \textbf{Input}: Normal training dataset $X$.\\
  \textbf{Parameter}: $\phi_{E}$, $\theta_{G}$, $\delta_{T}$ $\leftarrow$ Initialize network parameters \\
  \textbf{Output}: An encoder net $E$ and a generator net $G$.\par
  \begin{algorithmic}[1] 
  \FOR{$i \gets 1$ to $k$}
    \STATE $x\leftarrow$Random mini-batch from training dataset $X$.
    \STATE $\{\mu,\sigma\}\leftarrow$$E_{\phi}(x)$.
    \STATE $\{\mu_{T},\sigma_{T}\}\leftarrow$$T_{\delta}(\mu,\sigma)$.
    \STATE $z\leftarrow$Samples from $\mathcal { N } \left(   \mu  ,  \sigma ^ { 2 }  \right)$, 
    
    $z_{T}\leftarrow$Samples from $\mathcal { N } \left(   \mu_{T}  ,  \sigma_{T} ^ { 2 }  \right)$.
    \STATE $x_{r}\leftarrow$$G_{\theta}(z)$, $x_{Tr}\leftarrow$$G_{\theta}(z_{T})$.
    \STATE $\{\mu_{r},\sigma_{r}\}\leftarrow$$E_{\phi}(x_{r})$, $\{\mu_{Tr},\sigma_{Tr}\}\leftarrow$$E_{\phi}(x_{Tr})$.
    \STATE Calculating $\mathcal { L } _ { G }$ and $\mathcal { L } _ { T}$.
    \STATE $\{\theta_{G}, \delta_{T}\}\leftarrow\{\theta_{G}, \delta_{T}\}-\eta\nabla_{\theta,\delta} (\mathcal { L } _ { G } +\lambda\mathcal { L } _ { T})$.
    \STATE $x_{r}\leftarrow$$G_{\theta}(z)$, $x_{Tr}\leftarrow$$G_{\theta}(z_{T})$.
    \STATE $\{\mu_{r},\sigma_{r}\}\leftarrow$$E_{\phi}\left(detach(x_{r})\right)$, 
    
    $\{\mu_{Tr},\sigma_{Tr}\}\leftarrow$$E_{\phi}\left(detach(x_{Tr})\right)$.
    \STATE Calculating $\mathcal { L } _{E}$.
    \STATE $\phi_{E}\leftarrow\phi_{E}-\eta\nabla_{\phi} (\mathcal { L } _{E})$.
    \ENDFOR
  \end{algorithmic}
  
  \end{algorithm}
As described in Algorithm~\ref{alg:adVAE}, we train alternatively between the above two steps in a mini-batch iteration. These two steps are repeated until convergence. $detach(\cdot)$ indicates that the back propagation of the gradients is stopped at this point. 

In the first training step, the Gaussian transformer converts normal latent variables into anomalous latent variables. At the same time, the generator is trained to generate realistic-like samples when the latent variables are normal and to synthesize a low-quality reconstruction when they are not normal. It offers the generator the ability to distinguish between the normal and the anomalous latent variables. In the second training step, the encoder not only maps the samples to the prior latent distribution, but also attempts to distinguish between the real data $x$ and generated samples $x_{r}$. 

Importantly, we introduce the competition of $E$ and $G$ into our adVAE model by training alternatively between these two steps. Analogously to GANs, the generator is trained to fool the encoder in training step 1, and the encoder is encouraged to discriminate the samples generated by the generator in step 2. In addition to benefitting from adversarial alternative learning as in GANs, the encoder and generator models will also learn jointly for the given training data to maintain the advantages of VAEs.
\subsection{Anomaly Score}
\label{score}
As demonstrated in Figure~\ref{fig:adVAE} (c), only the generator and the encoder are used in the testing phase, as in a traditional VAE. 
Given a test data point $x\in \bm{R}^d$ as the input, the encoder estimates the parameters of the latent Gaussian variables $\mu$ and $\sigma$ as the output. Then, the reparameterization trick is used to sample $\bm{z}=\{z^{(1)},z^{(2)},...,z^{(L)}\}$ according to the latent distribution $\mathcal { N } \left(   \mu  ,  \sigma ^ { 2 }  \right)$ , i.e., $z^{(l)} = \mu + \sigma \odot \varepsilon^{(l)}$, where $\varepsilon\sim\mathcal { N }(0,I)$ and $l=1,2,...L$. $L$ is set to 1000 in this work and used to improve the robustness of adVAE's performance. The generator receives $z^{(l)}$ as the input and outputs the reconstruction $x_{r}^{(l)}\in \bm{R}^d$. 

The error between the inputs $x$ and their average reconstruction $\sum^{L}_{l=1}x_{r}^{(l)}$ reflects the deviation between the testing data and the normal data distribution learned by adVAE, such that the anomaly score of a mini-batch data $\bm{x} \in \bm{R}^{n\times d}$ ($n$ is the batch size) is defined as follows: 
\begin{linenomath}
\begin{equation}
    \label{equation:SPE1}
    \begin{aligned} 
    \bm{S}= & \bm{e}\bm{e}^T \\ =& \left(\bm{x}-\frac{1}{L}\sum^{L}_{l=1}\bm{x_{r}^{(l)}}\right)\left(\bm{x}-\frac{1}{L}\sum^{L}_{l=1}\bm{x_{r}^{(l)}}\right)^{T},
    \end{aligned} 
    \end{equation}
\begin{equation}
    \label{equation:SPE2}
    s=\{\bm{S}_{11},\bm{S}_{22},...,\bm{S}_{nn}\}
    \end{equation}
\end{linenomath}
with error matrix $\bm{e}\in \bm{R}^{n\times d}$ and squared error matrix $\bm{S}\in \bm{R}^{n\times n}$. $s\in \bm{R}^n$ is the anomaly scores vector of a mini-batch dataset $\bm{x}$. The dimension of $s$ is always equal to the batch size of $\bm{x}$. 
Data points with a high anomaly score are classified as anomalies. To determine whether an anomaly score is high enough, we need a decision threshold. In the next subsection, we will illustrate how to decide the threshold automatically by KDE \cite{DBLP:journals/csda/GramackiG17}.
\subsection{Decision Threshold}
\label{KDE}
Earlier VAE-based anomaly detection studies~\cite{an2015variational,xu2018unsupervised} often overlook the importance of threshold selection. Because we have no idea about what the values of reconstruction error actually represent, determining the threshold of reconstruction error is cumbersome. \diff{Some studies~\cite{an2015variational,8279425,suh2016echo} adjust the threshold by cross-validation. However, building a big enough validation set is a luxury in some cases, as anomalous samples are challenging to collect. Other attempts~\cite{xu2018unsupervised,DAGMM} simply report the best performance in the test dataset to evaluate models, which makes it difficult to reproduce the results in practical applications.} Thus, it is exceedingly important to let the anomaly detection model automatically determine the threshold. 

The KDE technique \cite{DBLP:journals/csda/GramackiG17} is used to determine the decision threshold \diff{in the case where only normal samples are provided.} Given the anomaly scores vector $s$ of the training dataset, KDE estimates the probability density function (PDF) $p(s)$ in a nonparametric way:
\begin{linenomath}
\begin{equation}
p(s)\approx \frac{1}{m h} \sum_{i=1}^{m} K\left(\frac{s-s_{i}}{h}\right)
\end{equation}
\end{linenomath}
where $m$ is the size of the training dataset, $\{s_i\}$, $i=1,2...,m$, is the training dataset's anomaly scores vector, $K(\cdot)$ is the kernel function, and $h$ is the bandwidth.

Among all kinds of kernel functions, radial basis functions (RBFs) are the most commonly used in density estimation. Therefore, a RBF kernel is used to estimate the PDF of the normal training dataset:
\begin{linenomath}
\begin{equation} 
    K_{RBF}(s ; h) \propto \exp \left(-\frac{s^{2}}{2 h^{2}}\right).
\end{equation}
\end{linenomath}

In practice, the choice of the bandwidth parameter $h$ has a significant influence on the effect of the KDE. To make a good choice of bandwidth parameters, Silverman~\cite{silverman2018density} derived a selection criterion: 
\begin{linenomath}
\begin{equation} 
    h=(\frac{m(d + 2)}{4})^{-\frac{1}{(d + 4)}}
\end{equation}
\end{linenomath}
where $m$ is the number of training data points and $d$ the number of dimensions.
After obtaining the PDF $p(s)$ of the training dataset by KDE, the cumulative distribution function (CDF) $F(s)$ can be obtained by equation~(\ref{equation:CDF}):
\begin{linenomath}
\begin{equation} 
    \label{equation:CDF}
F(s)=\int_{-\infty}^{s} p(s) d s
\end{equation}
\end{linenomath}
Given a significance level $\alpha\in[0,1]$ and a CDF, we can find a decision threshold $s_\alpha$ that satisfies $F(s_{\alpha})=1-\alpha$. In this case, there is at least $(1-\alpha)100\%$ probability that a sample with the anomaly score $s\geq s_{\alpha}$ is an outlier. Because KDE decides the threshold by estimating the interval of normal data's anomaly scores, it is clear that using KDE to decide the threshold is more objective and reasonable than simply relying on human experience. A higher significance level $\alpha$ leads to a lower missing alarm rate, which means that models have fewer chances to mislabel outliers as normal data. On the contrary, a lower $\alpha$ means a lower false alarm rate. Therefore, the choice of the significance level $\alpha$ is a tradeoff. The significance level $\alpha$ is recommended to be set to 0.1 for anomaly detection tasks.

\subsection{Detecting Outliers}
\begin{figure}[tb!]
    \centering
    \includegraphics[width=0.95\linewidth]{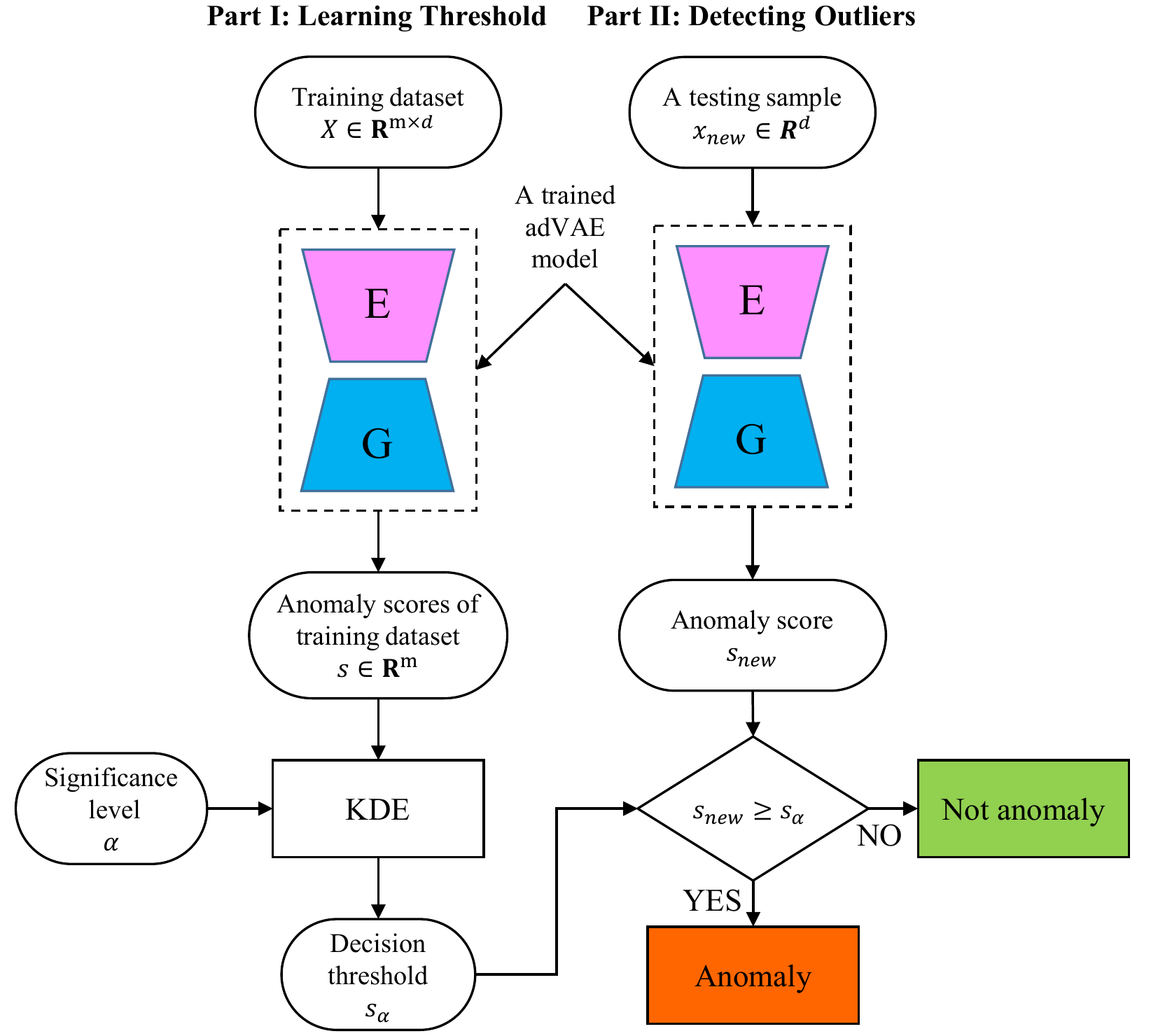}
    \caption{Work flow of using a trained adVAE model to choose threshold automatically and detect outliers.}
    \label{fig:flow-adVAE}
\end{figure}
\label{detect}
In this subsection, we summarize how to use a trained adVAE model and the KDE technique to learn the threshold from a training dataset and detect outliers in a testing dataset. As illustrated in Figure~\ref{fig:flow-adVAE}, this process is divided into two parts. 

Part \uppercase\expandafter{\romannumeral1} focuses on the training process. Given a training dataset matrix $X\in \bm{R}^{m\times d}$ consisting of normal samples, we can calculate their anomaly scores vector $s\in \bm{R}^m$ (described in subsection~\ref{score}), where $m$ is the size of training dataset and $d$ is the dimension of the data. As described in subsection~\ref{KDE}, the PDF $p(s)$ of the anomaly scores vector is obtained by KDE and the CDF $F(s)$ is obtained by $\int^{s}_{-\infty} p(s)ds$. Then, the decision threshold $s_{\alpha}$ can be determined from a given significance level $\alpha$ and CDF $F(s)$.

Part \uppercase\expandafter{\romannumeral2} is simple and easy to understand. The anomaly score $s_{new}$ of a new sample $x_{new}$ is calculated by adVAE. If $s_{new}\geq s_{\alpha}$, then $x_{new}$ is defined as an outlier. If not, then $x_{new}$ is regarded as a normal sample.  

\section{Experiments}
\subsection{Datasets}   
\begin{table}[tb!]
\caption{Summary of datasets used in experiments.}
\centering
\resizebox{\columnwidth}{!}{
\begin{tabular}{lrrr}
\toprule
Datasets & Size & Dim. & Outliers (percentage) \\ 
\midrule
Letter~\cite{DBLP:journals/tkdd/RayanaA16} & 1600 & 32 & 100 (6.2\%) \\ 
Cardio~\cite{DBLP:conf/sdm/SatheA16} & 1831 & 21 & 176 (9.6\%)\\
Satellite~\cite{DBLP:conf/icdm/LiuTZ08} & 5100 & 36 & 75 (1.5\%) \\ 
Optical~\cite{DBLP:journals/sigkdd/AggarwalS15} & 5216 & 64 & 150 (2.8\%) \\ 
Pen~\cite{DBLP:conf/icde/KellerMB12} & 6870 & 16 & 156 (2.3\%) \\  
\bottomrule
\end{tabular}
}
\label{tab:Datasets}
\end{table}
Most previous works used image datasets to test their anomaly detection models. To eliminate the impact of different convolutional structures and other image tricks on the test performance, we chose five publicly available and broadly used tabular anomaly detection datasets to evaluate our adVAE model. All the dataset characteristics are summarized in Table~\ref{tab:Datasets}. For each dataset, 80\% of the normal data were used for the training phase, and then the remaining 20\% and all the outliers were used for testing. More details about the datasets can be found in their references or our Github repository\footnote{\url{https://github.com/WangXuhongCN/adVAE}}.
\subsection{Evaluation Metric}
The anomaly detection community defines anomalous samples as positive and defines normal samples as negative, hence the anomaly detection tasks can also be regarded as a two-class classification problem with a large skew in class distribution. For the evaluation of a two-class classifier, the metrics are divided into two categories; one category is defined at a single threshold, and the other category is defined at all possible thresholds.

\textbf{Metrics at a single threshold.} Accuracy, precision, recall, and the F1 score are the common metrics to evaluate models performance at a single threshold. Because the class distribution is skew, accuracy is not a suitable metric for anomaly detection model evaluation. 

High precision means the fewer chances of misjudging normal data, and a high recall means the fewer chances of models missing alarming outliers. Even if models predict a normal sample as an outlier, people can still correct the judgment of the model through expert knowledge, because the anomalous samples are of small quantity. However, if models miss alarming outliers, we cannot find anomalous data in such a huge dataset. Thus, precision is not as crucial as recall. The F1 score is the harmonic average of precision and recall. Therefore, we adopt recall and the F1 score as the metrics for comparing at a single threshold.

\textbf{Metrics at all possible thresholds.} The anomaly detection community often uses receiver operator characteristic (ROC) and precision--recall (PR) curves, which aggregate over all possible decision thresholds, to evaluate the predictive performance of each method. When the class distribution is close to being uniform, ROC curves have many desirable properties. However, because anomaly detection tasks always have a large skew in the class distribution, PR curves give a more accurate picture of an algorithm's performance \cite{davis2006relationship}.

Rather than comparing curves, it is useful and clear to analyze the model performance quantitatively using a single number. Average precision (AP) and area under the ROC curve (AUC) are the common metrics to measure performance, with the former being preferred under class imbalance.

AP summarizes a PR curve by a sum of precisions at each threshold, multiplied by the increase in recall, which is a close approximation of the area under the PR curve: $AP=\sum _ { n } P _n \Delta R _n$, where $P _n$ is the precision at the $n^{th}$ threshold and $\Delta R_{n}$ is the increase in recall from the $n-1^{th}$ to the $n^{th}$ threshold. Because the PR curve is more useful than the ROC curve in anomaly detection, we recommend using AP as an evaluation metric for anomaly detection models rather than AUC.
In our experiments, recall, F1 score, AUC, and AP were used to evaluate the models performance.  
\subsection{Model Configurations}
\begin{figure}[tb!]
    \centering
    \includegraphics[width=0.95\linewidth]{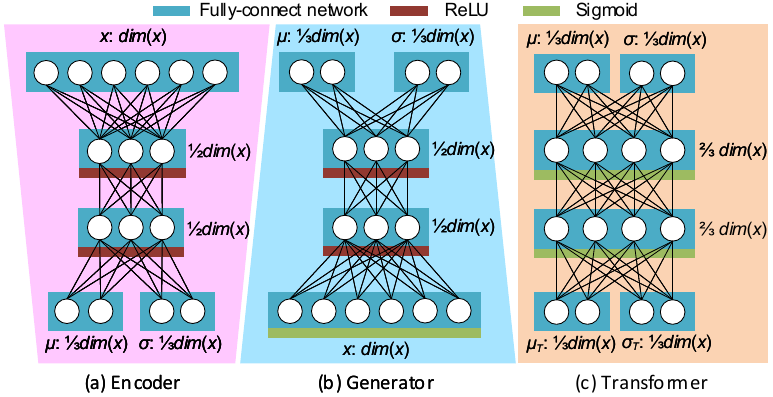}
    \caption{\diff{Network structure of adVAE. For the encoder, generator, and transformer networks, we adopted fully connected neural networks consisting of three hidden layers. For the individual dataset, we adjust the number of neurons according to the dimension of the training data $dim(x)$.}}
    \label{fig:network-conf}
\end{figure}
\diff{The network structures of adVAE used are summarized in Figure~\ref{fig:network-conf}, where the size of each layer was proportionate to the input dimension $dim(x)$. The Adam optimizer was used in all datasets with a learning rate of 0.001. We trained each model for a maximum 20000 mini-batch iterations with a batch size of 128. Kaiming uniform weight initialization was used for all three subnetworks.} 

\begin{table}[tb!]
    \caption{Summary of hyperparameters used in each dataset.}
    \centering
    \begin{tabular}{lrrrr}
    \toprule
    Datasets & $\lambda$ & $m_z$ & $m_x$ & $\gamma$ \\ 
    \midrule
    Letter~\cite{DBLP:journals/tkdd/RayanaA16} & 0.003 &40 &2&0.001  \\ 
    Cardio~\cite{DBLP:conf/sdm/SatheA16} & 0.1 & 20 & 2&0.001\\
    Satellite~\cite{DBLP:conf/icdm/LiuTZ08} & 0.01 & 40 & 2 &0.001 \\ 
    Optical~\cite{DBLP:journals/sigkdd/AggarwalS15} & 0.03 & 40 & 2 &0.001 \\ 
    Pen~\cite{DBLP:conf/icde/KellerMB12} & 0.01 & 20 & 2 &0.001 \\  
    \bottomrule
    \end{tabular}
    \label{tab:hyperparameter}
\end{table}
\diff{As shown in Table~\ref{tab:hyperparameter}, there are four hyperparameters in adVAE model. $\lambda$ is derived from plain VAE, and the other three new parameters are used to maintain the balance between the original training objective and the additional discrimination objective. The larger the $\gamma$ or $m_z$, the larger the proportion of the encoder discrimination objective in the total loss. The larger the $m_x$, the larger the proportion of the generator discrimination objective.}

 
The KLD margin $m_{z}$ can be selected to be slightly larger than the training KLD value of plain VAEs, and the MSE margin $m_{x}$ was set equal to 2 for each dataset. \diff{Because $\gamma$ and $m_z$ have similar effects, $\gamma$ is always suggested to be fixed as 0.001. Adjusting the parameters becomes easier, because we can simply consider tuning $m_{x}$ and $m_{z}$.} Note that the structure and hyperparameters used in this study proved to be sufficient for our applications, although they can still be improved. 

\subsection{Compared Methods}
\begin{figure}[tb!]
    \centering
    \includegraphics[width=0.95\linewidth]{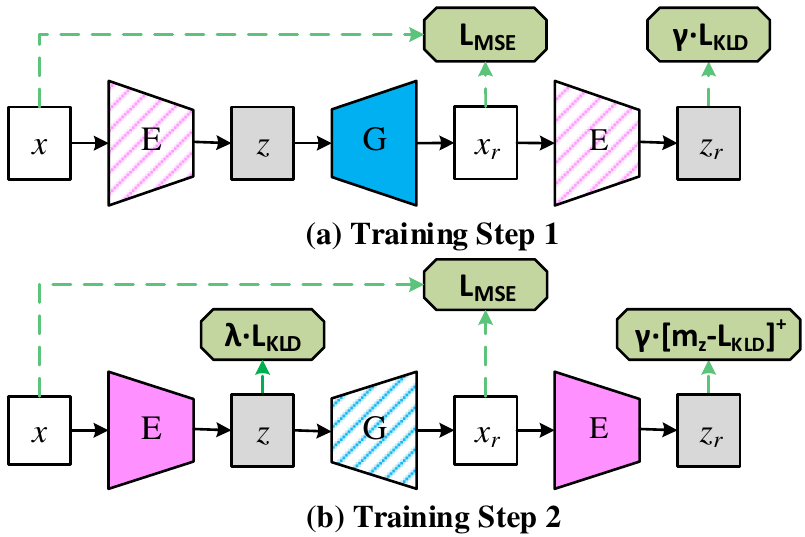}
    \caption{\diff{Structure of E-adVAE. As adVAE is formed by adding a discrimination objective to both the generator and the encoder of plain VAE, we also propose an ablation model E-adVAE, in which only the encoder has the discrimination objective.}}
    \label{fig:E-adVAE}
\end{figure}

\begin{figure}[tb!]
    \centering
    \includegraphics[width=0.95\linewidth]{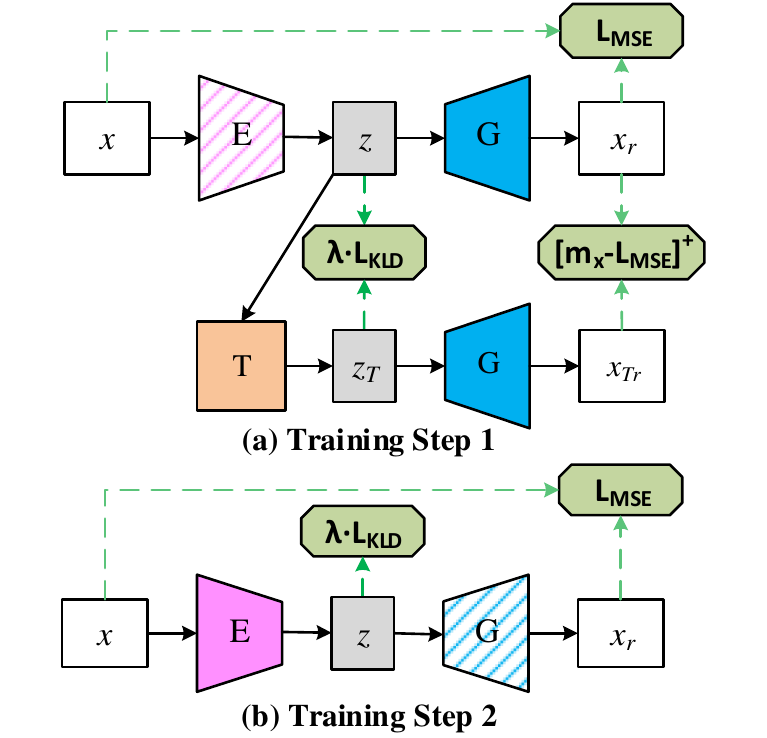}
    \caption{\diff{Structure of G-adVAE. This is the other ablation model G-adVAE: only the generator has the discrimination objective.}}
    \label{fig:G-adVAE}
\end{figure}

\diff{We compare adVAE with the eleven most advanced outlier detection algorithms, which include both traditional and deep-learning-based methods. To obtain a convincing conclusion, the hyperparameters of competing methods are searched in a range of optional values. They can be divided into seven categories:
(\romannumeral1) A boundary-based approach, OCSVM~\cite{OCSVM}. It is a widely used semisupervised outlier detection algorithm, and we use a RBF kernel in all tasks. Because OCSVM needs an important parameter $\nu$, its value will be searched in the range \{0.01, 0.05, 0.1, 0.15, 0.2\}.
(\romannumeral2) An ensemble-learning-based approach, IForest~\cite{Iforest}. The hyperparameter to be tuned is the number of decision trees $n_t$, which is chosen from the set \{50, 100, 200, 300\}. 
(\romannumeral3) A probabilistic approach, ABOD~\cite{ABOD}.  
(\romannumeral4) Distance-based approaches, SOD~\cite{SOD} and HBOS~\cite{HBOS}. Because the performance of ABOD and SOD will be dramatically affected by the size of the neighborhood, we tune it in the range of \{5, 8, 16, 24, 40\}. For HBOS, the number of bins is chosen from \{5, 10, 20, 30, 50\}.
(\romannumeral5) Three GAN-based models, GAN~\cite{goodfellow2014generative}, MOGAAL~\cite{8668550}, and WGAN-GP~\cite{DBLP:conf/nips/GulrajaniAADC17}. GAN shares the same network structure as adVAE, except that the output layer of the discriminator is one-dimensional, and the configurations of MOGAAL\footnote{\url{https://github.com/leibinghe/GAAL-based-outlier-detection}} refer to its official open source code. \revise{Based on the network structure of GAN, WGAN-GP removes the output activation function of the generator.}
(\romannumeral6) Three reconstruction-based methods, AE~\cite{an2015variational}, DAGMM~\cite{DAGMM}, and VAE~\cite{an2015variational}. The AE, VAE, and DAGMM share the same network structure as adVAE, except that the latent dimension of DAGMM\footnote{\url{https://github.com/danieltan07/dagmm}} is required to be 1. VAE shares the same hyperparameter $\lambda$ as adVAE. 
(\romannumeral7) Two ablation models of the proposed adVAE, named E-adVAE and G-adVAE, are also compared to demonstrate how the two discrimination objectives affect the performance. Figure~\ref{fig:E-adVAE} and \ref{fig:G-adVAE} show the architecture of our two ablation models, respectively. E-adVAE, G-adVAE, and adVAE share the same parameters and network structure.
}

\diff{All deep-learning methods are implemented in pytorch, and share the same optimizer, learning rate, batch size, and iteration times as adVAE, except that the parameters of DAGMM and MOGAAL refer to their author's recommendation. All traditional methods are implemented on a common outlier detection framework PyOD~\cite{zhao2019pyod}.}
\subsection{Results and Discussion}
\begin{table*}[htb!]
\caption{AP and AUC comparisons between the baselines and adVAE. AP is a better metric than AUC in anomaly detection. Note that the best result is typeset in \textbf{bold} and the second best in \textit{italic} typeface. The dataset names are represented as their first three letters and Avg. represents the average.}
\centering
\resizebox{\textwidth}{!}{
\begin{tabular}{lcccccccccccc}  
    \toprule
    \multirow{2}*{Methods} & \multicolumn{6}{c}{Average Precision (AP)}& \multicolumn{6}{c}{Area Under the ROC Curve (AUC)}  \\ 
    \cmidrule(lr){2-7}   \cmidrule(lr){8-13}
                                & Let & Car & Sat & Opt & Pen & Avg.              & Let & Car & Sat & Opt & Pen & Avg.\\ 
    \cmidrule(lr){1-7}    \cmidrule(lr){8-13}
    OCSVM~\cite{OCSVM}          & 0.306 & \textit{0.947} & 0.718 & 0.157 & 0.645 & 0.555         & 0.514 & \textbf{0.975} & 0.893 & 0.613 & 0.936 & 0.786 \\ 
    IForest~\cite{Iforest}      & 0.330 & 0.922 & 0.784 & 0.353 & 0.798 & 0.637         & 0.624 & 0.949 & 0.945 & 0.859 & 0.978 & 0.871 \\ 
    ABOD~\cite{ABOD}             & \textit{0.769} & 0.922 & \textbf{0.845} & 0.762 & \textit{0.917} & 0.843         & \textit{0.915} & 0.948 & \textbf{0.972} & 0.969 & \textit{0.994} & 0.960 \\ 
    HBOS~\cite{HBOS}            & 0.312 & 0.851 & 0.746 & 0.463 & 0.641 & 0.603        & 0.619 & 0.899 & 0.915 & 0.863 & 0.942 & 0.848\\
    SOD~\cite{SOD}              & 0.542 & 0.693 & 0.494 & 0.105 & 0.115 & 0.390        & 0.792 & 0.764 & 0.915 &0.405&0.473& 0.670\\
    GAN~\cite{goodfellow2014generative}          & 0.421 & 0.697 & 0.427 & 0.417 & 0.362 & 0.465         & 0.653 & 0.618 & 0.776 &0.743 & 0.779 & 0.714 \\ 
    MOGAAL~\cite{8668550}       & 0.428 & 0.730 & 0.797 & 0.588 & 0.308 & 0.570        & 0.714 & 0.792 & 0.971 & 0.912 & 0.838 & 0.845 \\ 
    WGAN-GP~\cite{DBLP:conf/nips/GulrajaniAADC17}          & 0.497 & 0.834 & 0.765 & 0.596 & 0.589 & 0.656         & 0.722 & 0.847 & 0.920 &0.895 & 0.951 & 0.867 \\
    AE~\cite{an2015variational} & 0.693 & 0.798 & 0.764 & 0.860 & 0.682 & 0.759        & 0.897 & 0.840 & 0.950 & 0.990 & 0.962 & 0.928 \\
    DAGMM~\cite{DAGMM}          & 0.318 & 0.789 & 0.353 & 0.125 & 0.370 & 0.391        & 0.564 & 0.847 & 0.714 & 0.465 & 0.826 & 0.683 \\
    VAE~\cite{an2015variational}& 0.724 & 0.803 & 0.766 & 0.875 & 0.625 & 0.759        & 0.911 & 0.840 & 0.962 & 0.992 & 0.959 & 0.933 \\
    \cmidrule(lr){1-7}    \cmidrule(lr){8-13}
    G-adVAE                     & 0.723 & 0.922 & 0.748 & \textit{0.885} & 0.853 & 0.826       & 0.908 & 0.956 & 0.957 & \textit{0.993} & 0.988 & 0.960 \\
    E-adVAE                     & \textit{0.769} & 0.933 & \textit{0.801} & 0.857 & \textbf{0.926} & \textit{0.857}        & 0.911 & 0.961 & 0.968 & 0.989 & \textbf{0.996} & \textit{0.965} \\
    adVAE                       & \textbf{0.779} & \textbf{0.951} & 0.792 & \textbf{0.957} & 0.880 & \textbf{0.872}        & \textbf{0.921} & \textit{0.966} & \textit{0.970} & \textbf{0.996} & 0.993 & \textbf{0.969} \\
    \bottomrule
\end{tabular}
}
\label{tab:APAUC}
\end{table*}
\begin{table*}[htb!]
\caption{Recall and F1 score comparisons between the baselines and adVAE. For all reconstruction-based methods (AE, VAE, DAGMM, G-adVAE, E-adVAE, and adVAE), the significance level $\alpha$ was set to 0.1. Note that the best result is typeset in \textbf{bold} and the second best in \textit{italic} typeface. The dataset names are represented as their first three letters and Avg. represents the average.}
\centering
\resizebox{\textwidth}{!}{
\begin{tabular}{lcccccccccccc}  
\toprule
\multirow{2}*{Methods} & \multicolumn{6}{c}{Recall}& \multicolumn{6}{c}{F1 Score}  \\ 
\cmidrule(lr){2-7}   \cmidrule(lr){8-13}
                            & Let & Car & Sat & Opt & Pen & Avg.                 & Let & Car & Sat & Opt & Pen & Avg.\\ 
\midrule
OCSVM~\cite{OCSVM}          & 0.140 & \textit{0.977} & 0.773 & 0.093 & 0.795 & 0.556         & 0.201 & \textbf{0.898} & 0.439 & 0.104 & 0.608 & 0.450 \\ 
IForest~\cite{Iforest}      & 0.160 & 0.830 & 0.827 & 0.407 & 0.987 & 0.642         & 0.215 & \textit{0.816} & 0.539 & 0.401 & 0.700 & 0.534 \\ 
ABOD~\cite{ABOD}            & 0.750 & \textbf{0.983} & \textit{0.893} & 0.940 & \textbf{1.000} & 0.913         & \textbf{0.739} & 0.647 & \textbf{0.657} & 0.709 & 0.674 & 0.685 \\ 
HBOS~\cite{HBOS}            & 0.200 & 0.625 & 0.853 & 0.587 & 0.827 & 0.618         & 0.256 & 0.703 & 0.490 & 0.507 & 0.620 & 0.515\\
SOD~\cite{SOD}              & 0.670 & 0.449 & 0.800 & 0.113 & 0.224 & 0.451         & 0.563 & 0.583 & 0.441 &0.089&0.131& 0.361\\
GAN~\cite{goodfellow2014generative}          & 0.480 & 0.903 & 0.747 & 0.713 & 0.692 & 0.707         & 0.425 & 0.500 & 0.226 &0.450 & 0.206 & 0.361 \\ 
MOGAAL~\cite{8668550}       & 0.180 & 0.455 & 0.493 & 0.367 & 0.276 & 0.354        & 0.277 & 0.578 & 0.627 & 0.455 & 0.319 & 0.451 \\ 
WGAN-GP~\cite{DBLP:conf/nips/GulrajaniAADC17}           & 0.490 & 0.824 & 0.813 & 0.707 & 0.955 & 0.758         & 0.500 & 0.700 & 0.419 &0.542 & 0.623 & 0.557 \\
AE~\cite{an2015variational} & 0.750 & 0.830 & 0.840 & 0.940 & 0.891 & 0.850        & 0.685 & 0.661 & 0.589 & 0.673 & 0.670 & 0.655 \\
DAGMM~\cite{DAGMM}          & 0.310 & 0.756 & 0.467 & 0.107 & 0.628 & 0.453        & 0.315 & 0.731 & 0.237 & 0.114 & 0.476 & 0.374 \\
VAE~\cite{an2015variational}& 0.760 & 0.801 & \textit{0.893} & \textbf{1.000} & 0.885 & 0.868        & 0.714 & 0.647 & 0.568 & \textbf{0.744} & 0.670 & 0.668 \\
\midrule
G-adVAE                     & 0.690 & 0.960 & 0.787 & \textbf{1.000} & \textbf{1.000} & 0.887        & 0.717 & 0.824 & 0.549 & 0.718 & \textbf{0.721} & 0.706 \\
E-adVAE                     & \textit{0.770} & 0.943 & \textbf{0.907} & \textbf{1.000} & \textbf{1.000} & \textbf{0.924}        & 0.691 & 0.804 & 0.618 & \textit{0.732} & \textit{0.717} & \textit{0.712} \\
adVAE                       & \textbf{0.780} & 0.938 & 0.853 & \textbf{1.000} & \textbf{1.000} & \textit{0.914}        & \textit{0.729} & 0.805 & \textit{0.631} & 0.718 & 0.704 & \textbf{0.717} \\
\bottomrule
\end{tabular}
}
\label{tab:RecallF1}
\end{table*}

Table~\ref{tab:APAUC} shows the experimental results for the AP and AUC metrics, and Table~\ref{tab:RecallF1} indicates the results for the recall and F1 score. According to the experimental results, the adVAE model achieved 11 best and 5 second-best results from 24 comparisons. Therefore, adVAE is significantly better than other compared methods. 

Among all the experimental results, the performance of VAE is generally slightly better than AE, because VAE has a KLD regularization of the encoder, which proves the importance of regularization in AE-based anomaly detection algorithms.

We customize plain VAE by the proposed self-adversarial mechanism, reaching the highest AP improvement of 0.255 (40.8\%). Compared to its base technique VAE, on average, adVAE improved by 0.113 (14.9\%) in AP, 0.036 (3.9\%) in AUC, 0.046 (5.3\%) in recall, and 0.049 (7.3\%) in F1 score. The outstanding results of adVAE prove the superiority of the proposed self-adversarial mechanism, which helps deep generative models to play a greater role in anomaly detection research and applications.

\diff{OCSVM} performed well in the cardio, satellite, and pen datasets, but achieved horrible results in the letter and optical datasets. The results indicate that the conventional method OCSVM is not stable enough, leading the AP gap from adVAE to reach an astonishing 0.8 in the optical dataset. \diff{Actually, all traditional anomaly detection methods (e.g., IForest, HBOS, and SOD) have the same problem as OCSVM: They perform well on some datasets, but they perform extremely poorly on other datasets. ABOD's performance is more stable than others, but it still suffers from the above problem.}

The reason for this is that conventional methods often have a strict data assumption. If training data are high-dimensional or not consistent with the data assumption, their performance may be significantly impaired. Because neural-network-based methods do not need a strict data assumption and are suitable for big and high-dimensional data, they have more stable performance on diverse datasets with different data distributions.

In our experiments, both VAE and AE outperformed the GAN-based methods in most datasets. The results are analyzed as follows:
Because the mode collapse issue of GANs cannot be characterized in a training loss curve, people usually monitor the generated images during the training phase. However, for tabular datasets, it is difficult for us to ensure that the generated samples are diverse enough. A GAN with mode collapse would mislabel normal data as outliers and cause a high false positive rate, because GAN learns an incomplete normal data distribution in this case. \diff{To better recover the distribution of normal data, MOGAAL suggests stopping optimizing the generator before convergence and expanding the network structure from a single generator to multiple generators with different objectives. However, a nonconverged model brings more uncertainty to experimental results, because it is impossible to know what the model has learned precisely. According to the results, it does have a performance improvement over plain GAN, but still has a performance gap with AE-based methods. \revise{WGAN-GP tries to avoid mode collapsing, which leads to better results of all four metrics than plain GAN. However, it still results in worse performance than encoder-decoder-based methods.} GAN is better at generating high-quality samples rather than learning the complete probability distribution of the training dataset. Thus GAN-based methods do not perform well for outlier detection.}

\diff{For three reconstruction-based neural network models, there is no significant performance gap between AE and VAE, whereas DAGMM is worse than them in our tests. This is explained by the fact that the latent dimension of DAGMM is required to be set as 1, which severely limits the ability of the model to summarize datasets. The results also indicate this explanation: DAGMM only achieves good results in the cardio dataset, which has the lowest data size and dimensions. The larger the size and dimensions of the dataset, the larger performance gap between DAGMM and AE.}

In conclusion, the AE and VAE methods are better than most of the other anomaly detection methods according to our tests. Moreover, adVAE learns to detect outliers through the self-adversarial mechanism, which further enhances the anomaly detection capability and achieves state-of-the-art performance. We believe that the proposed self-adversarial mechanism has immense potentiality, and it deserves to be extended to other pattern recognition tasks.
\subsection{Ablation Study}
\begin{figure}[tb!]
    \centering
    \includegraphics[width=0.95\linewidth]{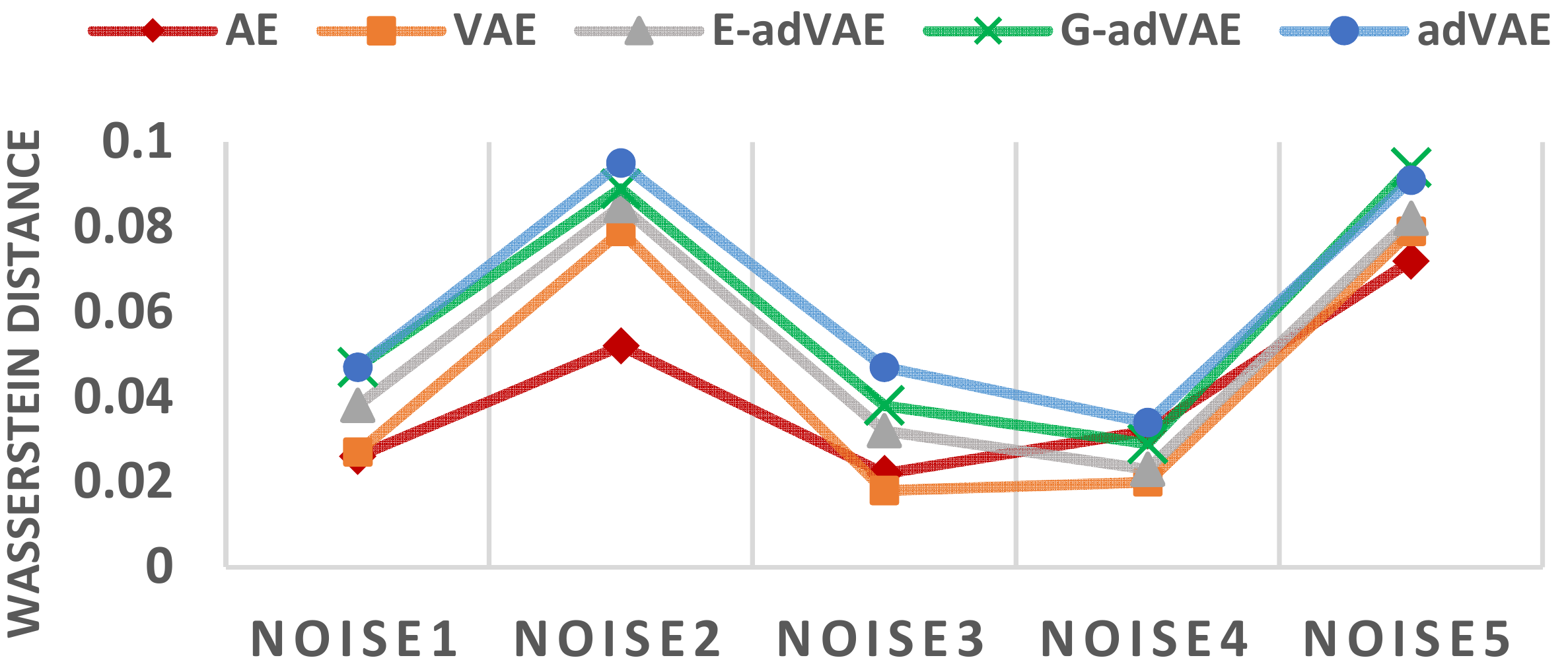}
    \caption{\diff{To independently verify the sensitivity of the generator to anomalies, we add different noises to the normal latent variables. Noise 1: adding a uniform distribution $\mathcal{U}(0,1)$; Noise 2: adding a Gaussian distribution $\mathcal { N }(0,1)$; Noise 3: adding a constant value 0.5; Noise 4: multiplying the last half dimensions of the latent vector by 0.5; Noise 5: setting the first half dimensions of the latent vector to 0. The larger the Wasserstein distance, the better the anomaly detection capability. The generator of adVAE achieves the best discrimination performance, and G-adVAE's generator is the second best. Benefiting from the adversarial training between $E$ and $G$, the detection performance of the E-adVAE's generator is also improved in comparison to plain VAE.}}
    \label{fig:G-power}
\end{figure}
\begin{figure*}[t!]
    \centering
    \subfigure[VAE]
    {
        \label{fig:VAE-letter-zspace}
        \includegraphics[width=0.22\textwidth]{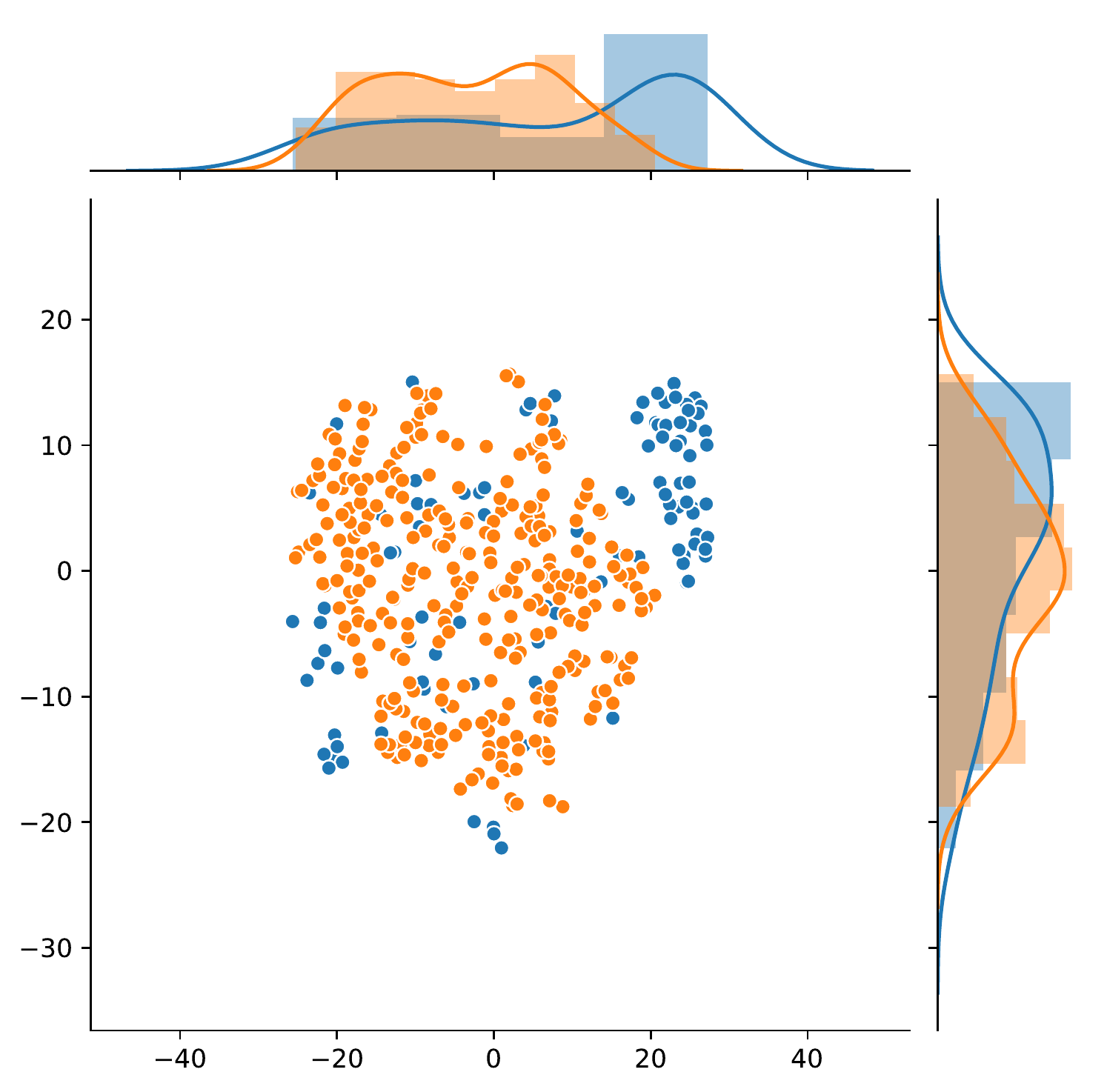}
    }
    \subfigure[G-adVAE]
    {
        \label{fig:G-adVAE-letter-zspace}
        \includegraphics[width=0.22\textwidth]{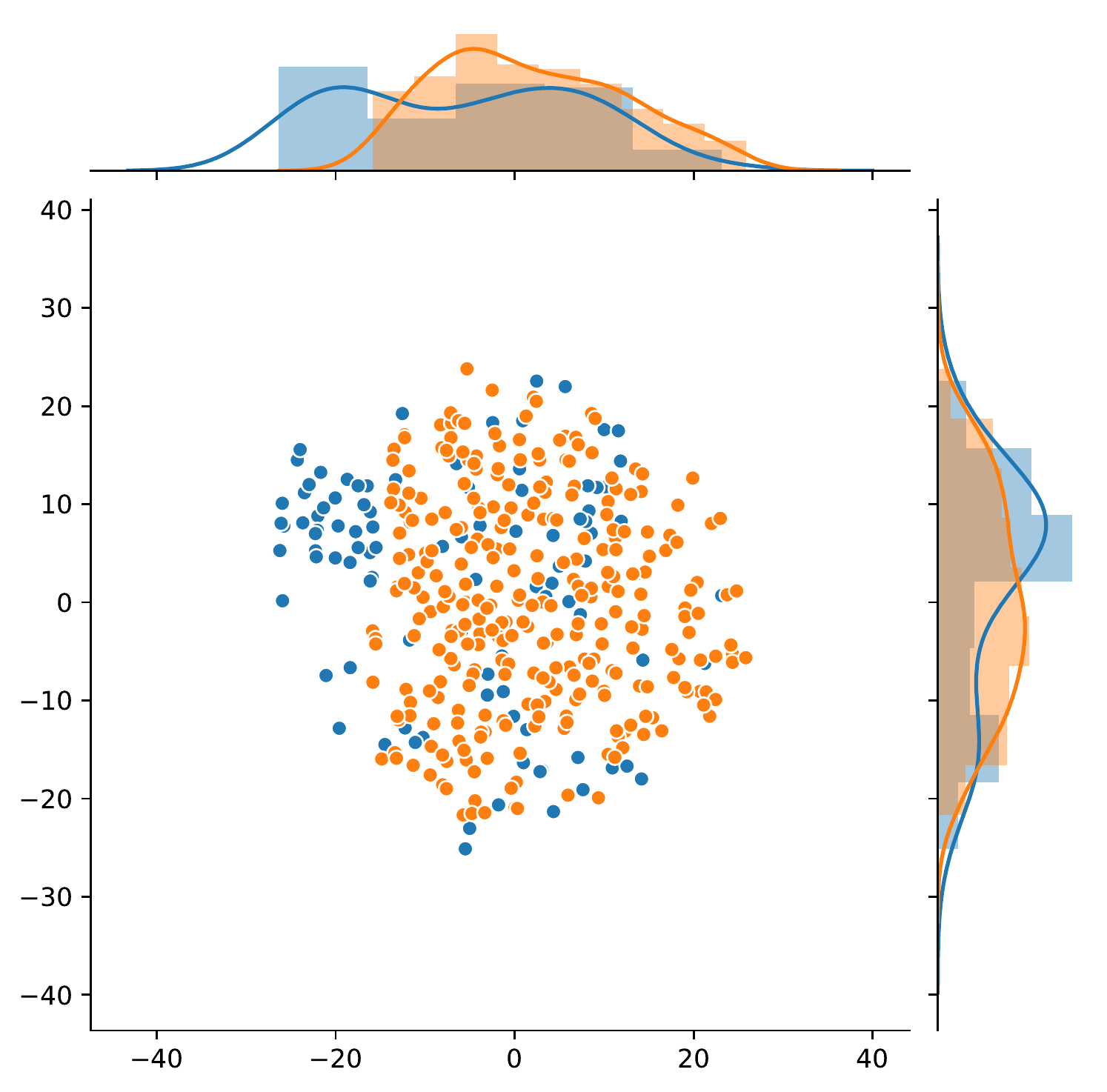}
    }
    \subfigure[E-adVAE]
    {
        \label{fig:E-adVAE-letter-zspace}
        \includegraphics[width=0.22\textwidth]{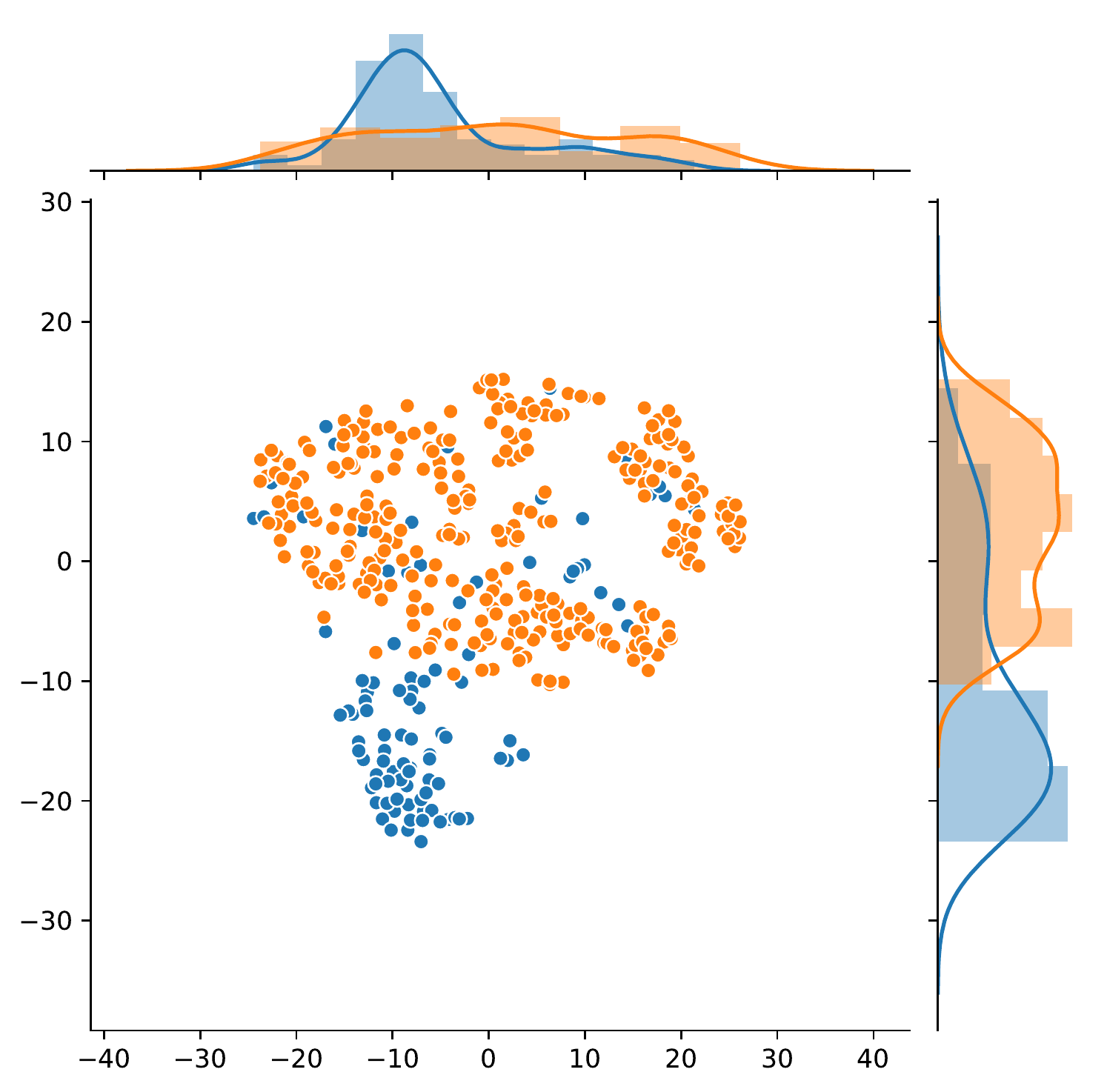}
    }
    \subfigure[adVAE]
    {
        \label{fig:adVAE-letter-zspace}
        \includegraphics[width=0.22\textwidth]{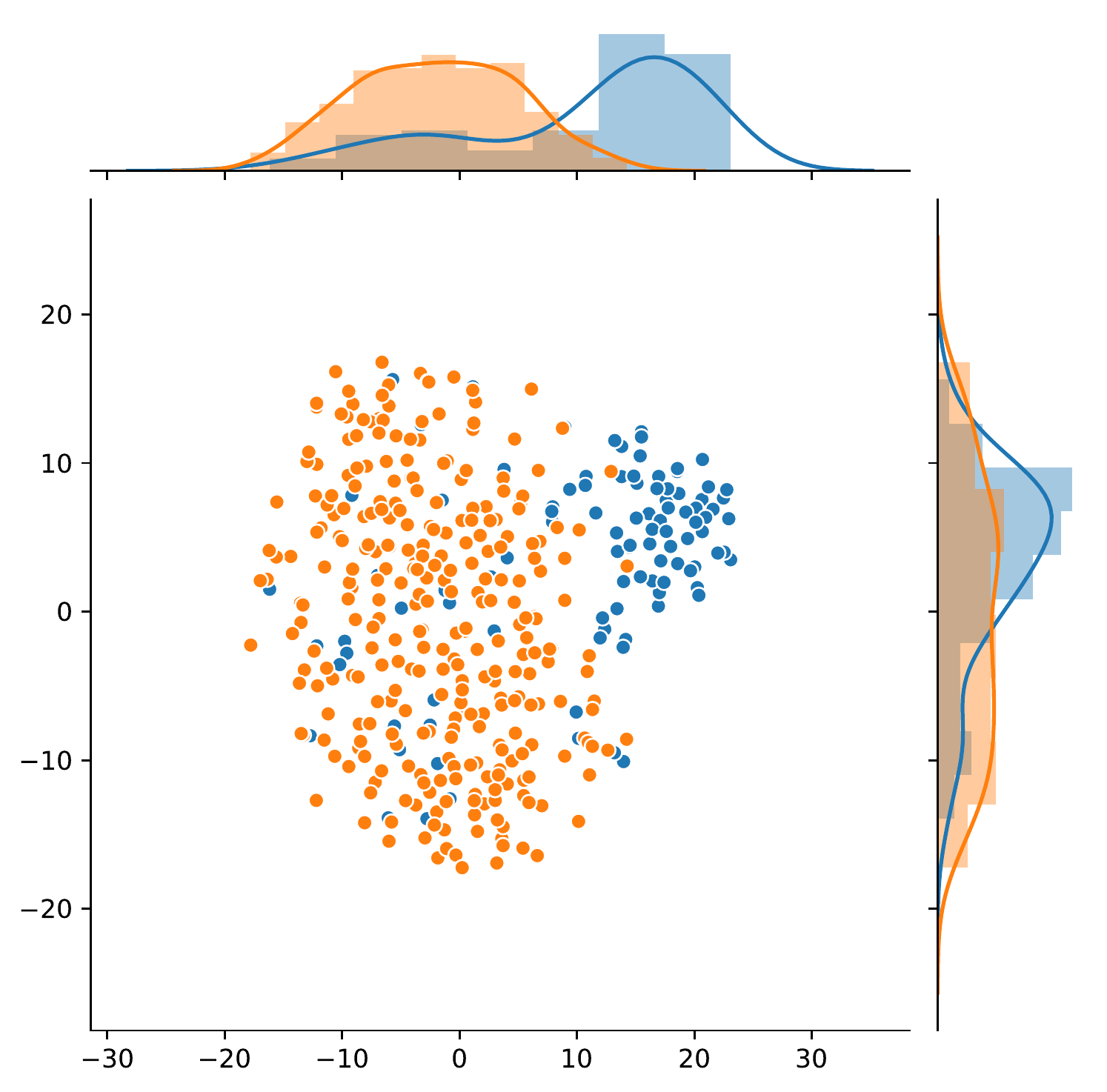}
    }
    \caption{T-SNE visualization of the latent space in the letter dataset. The orange points represent the latent variables encoded from normal data, and the blue points are encoded from anomalous data. The KDE curves at the top indicate that adVAE's latent space overlaps are smaller than VAE's, which means that the encoder of adVAE has a better ability to distinguish the normal data from the outliers. \diff{The visualization plot of G-adVAE does not show significant advantages compared to plain VAE, and E-adVAE's plot proves that the encoder does benefit from the discrimination objective.} 
    }
    \label{fig:zspace}
\end{figure*}
\begin{figure*}[t!]
    \centering
    \subfigure[$\lambda$]
    {
        \label{fig:lambda}
        \includegraphics[width=0.4\linewidth]{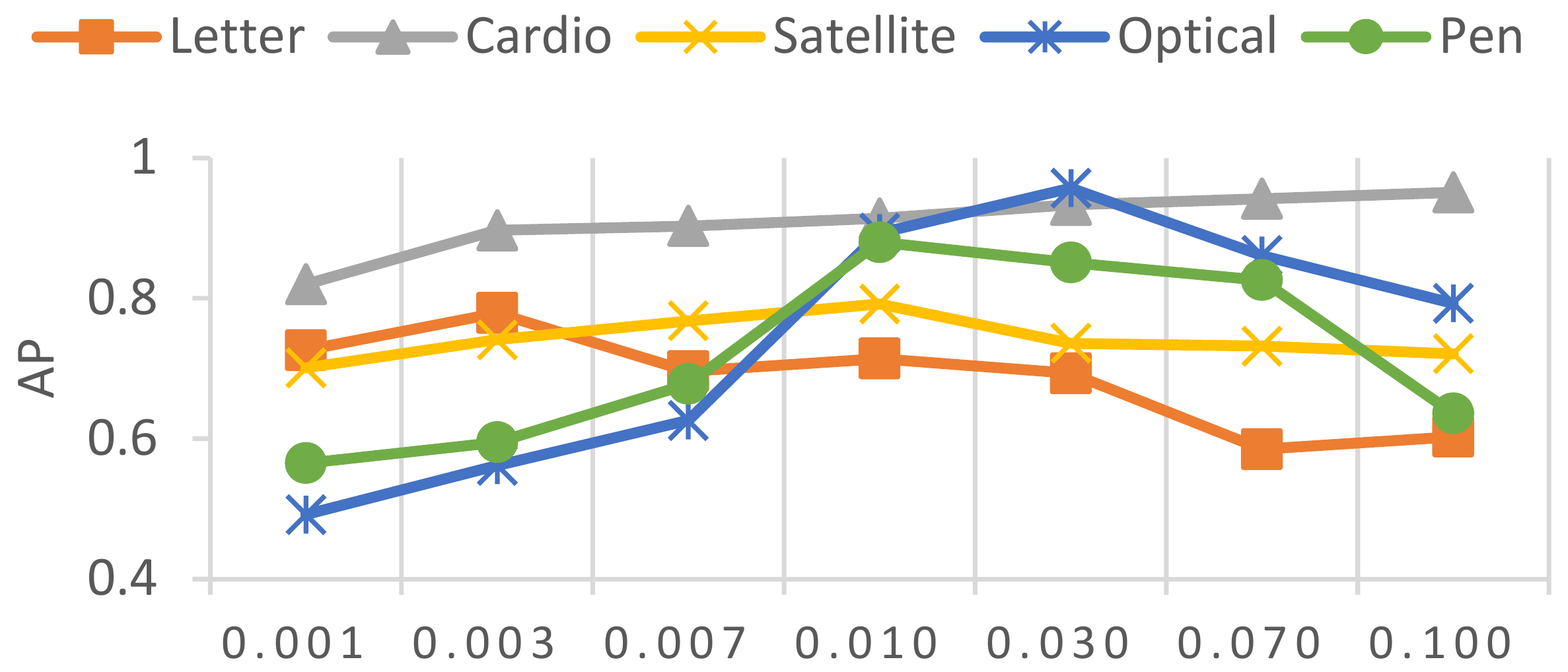}
    }
    \subfigure[$\gamma$]
    {
        \label{fig:gamma}
        \includegraphics[width=0.4\linewidth]{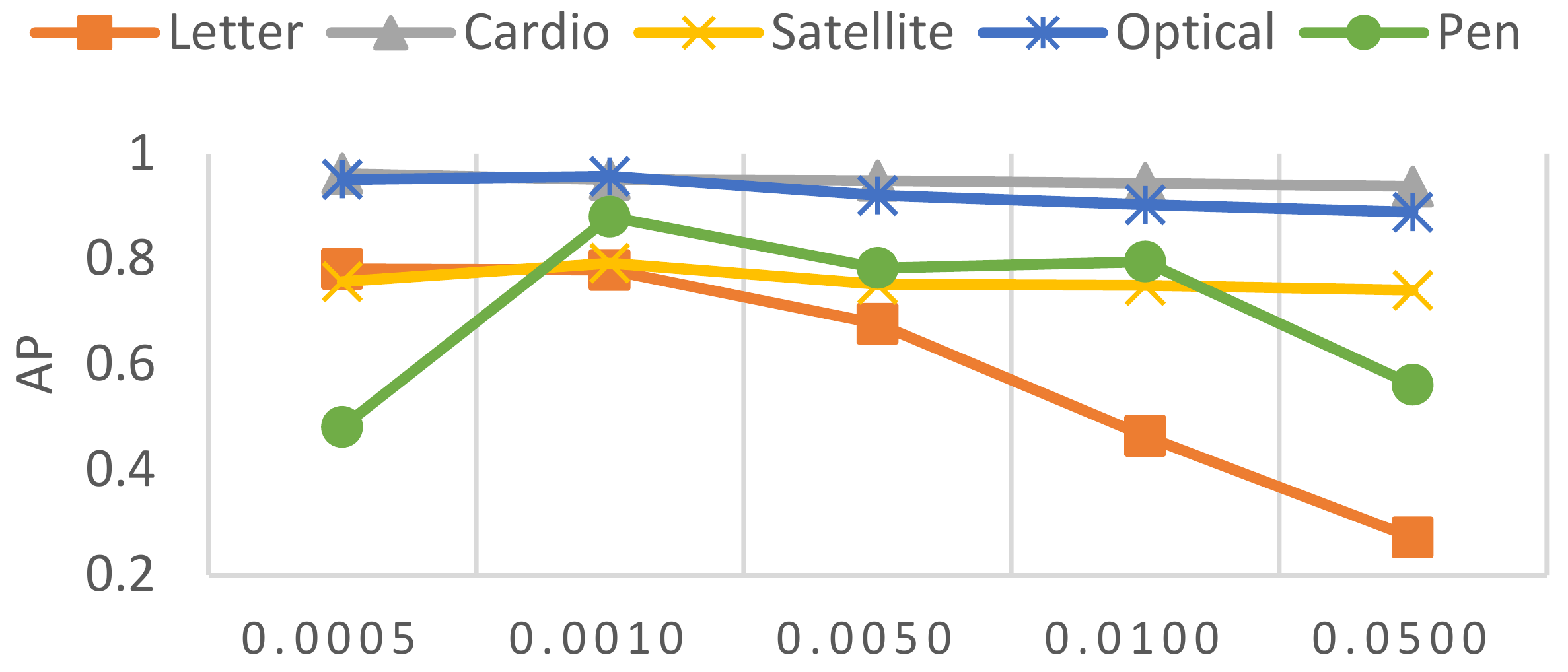}
    }
    \subfigure[$m_z$]
    {
        \label{fig:KLD-margin}
        \includegraphics[width=0.4\linewidth]{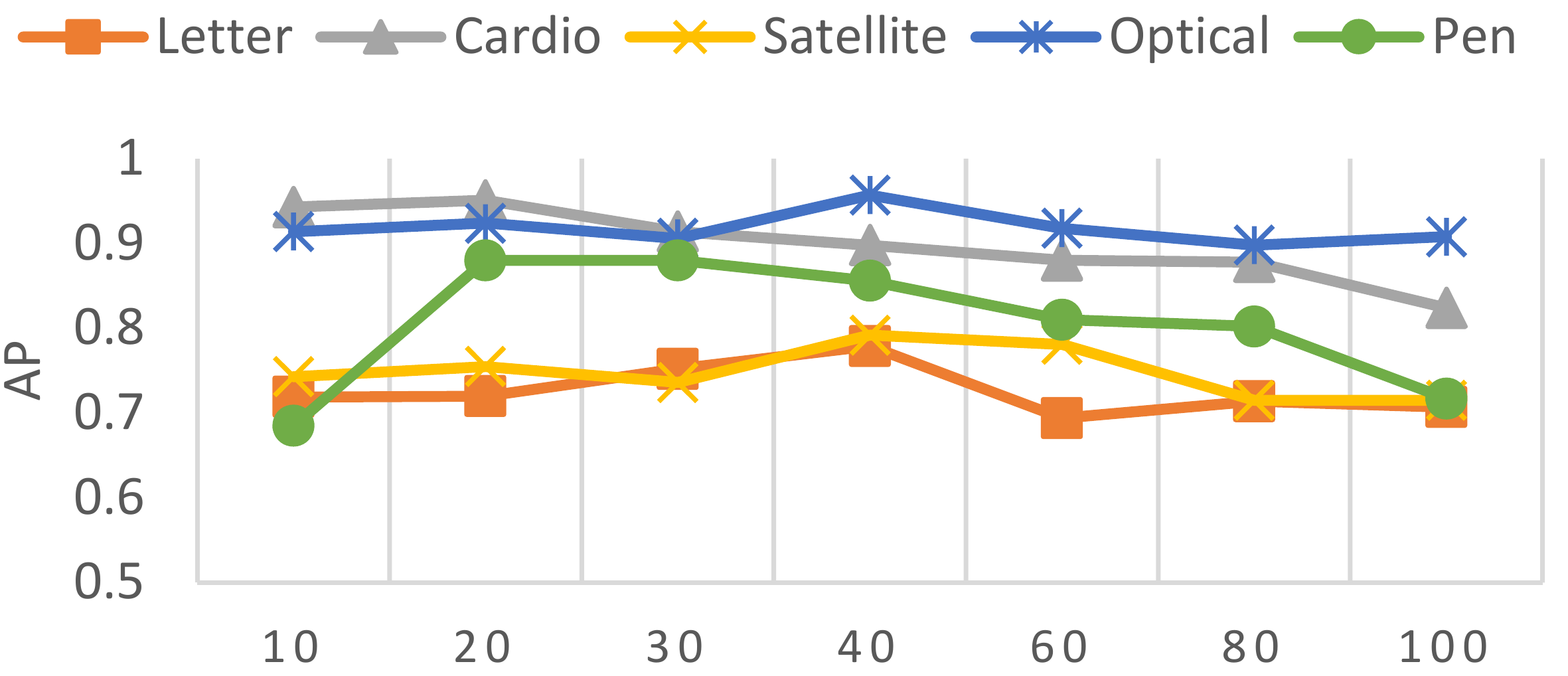}
    }
    \subfigure[$m_x$]
    {
        \label{fig:MSE-margin}
        \includegraphics[width=0.4\linewidth]{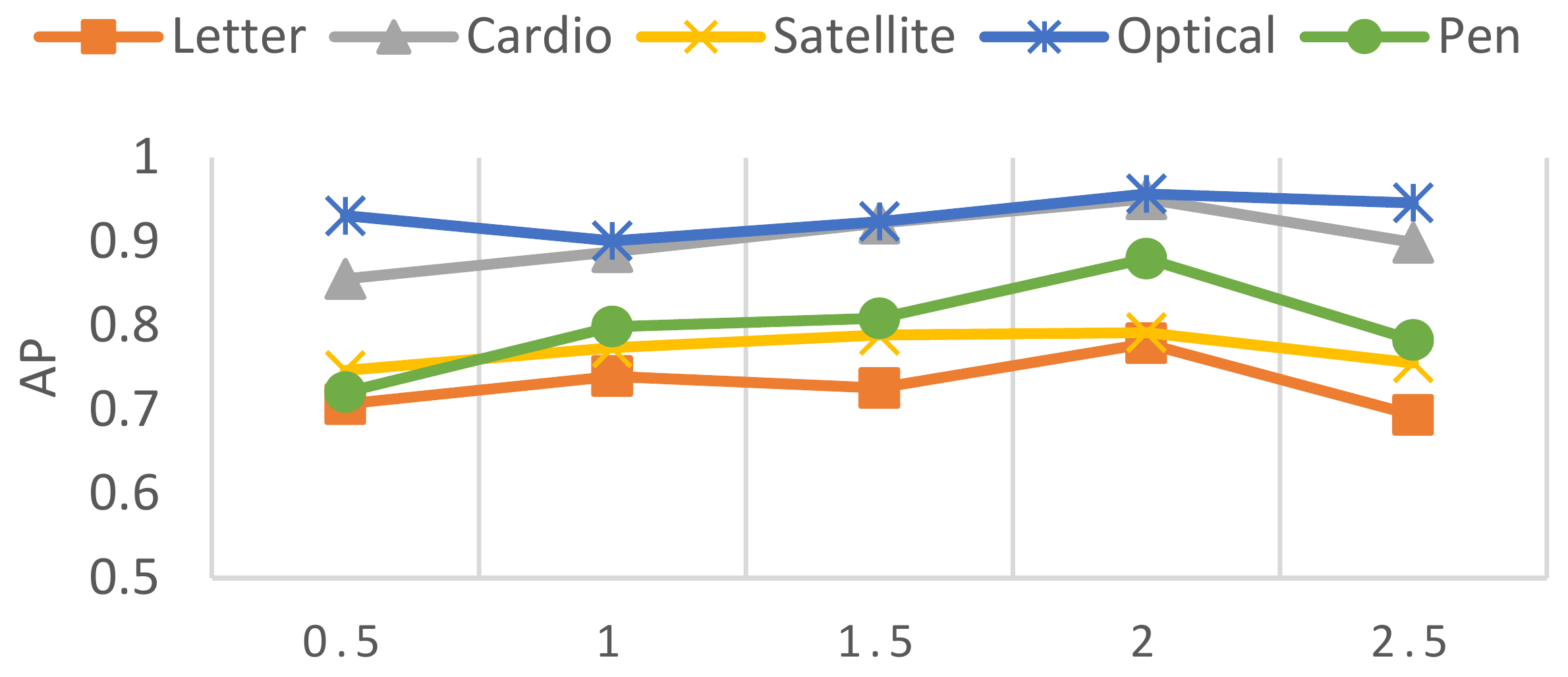}
    }
    \subfigure[Network Structure]
    {
        \label{fig:network-structure}
        \includegraphics[width=0.4\linewidth]{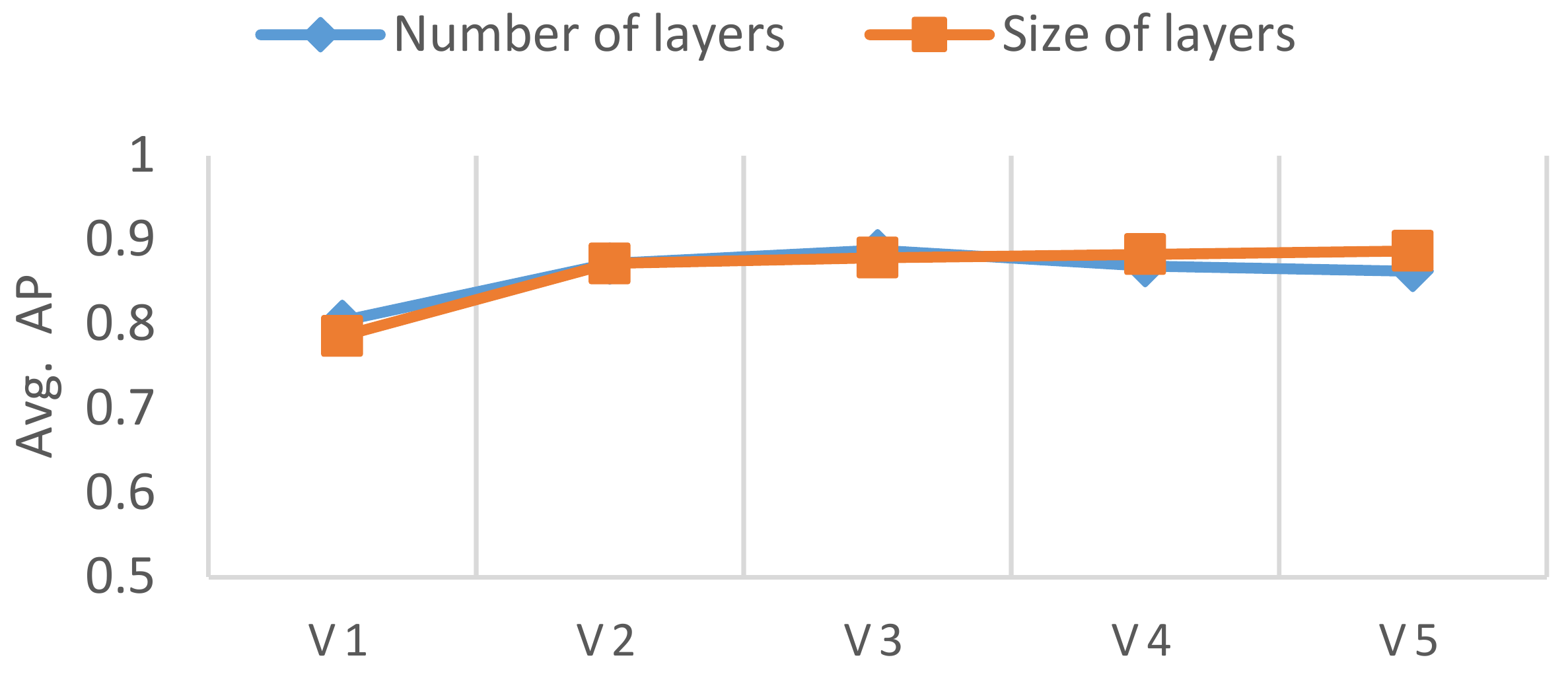}
    }
    \subfigure[Training Dataset Contamination Ratio]
    {
        \label{fig:add-outliers}
        \includegraphics[width=0.4\linewidth]{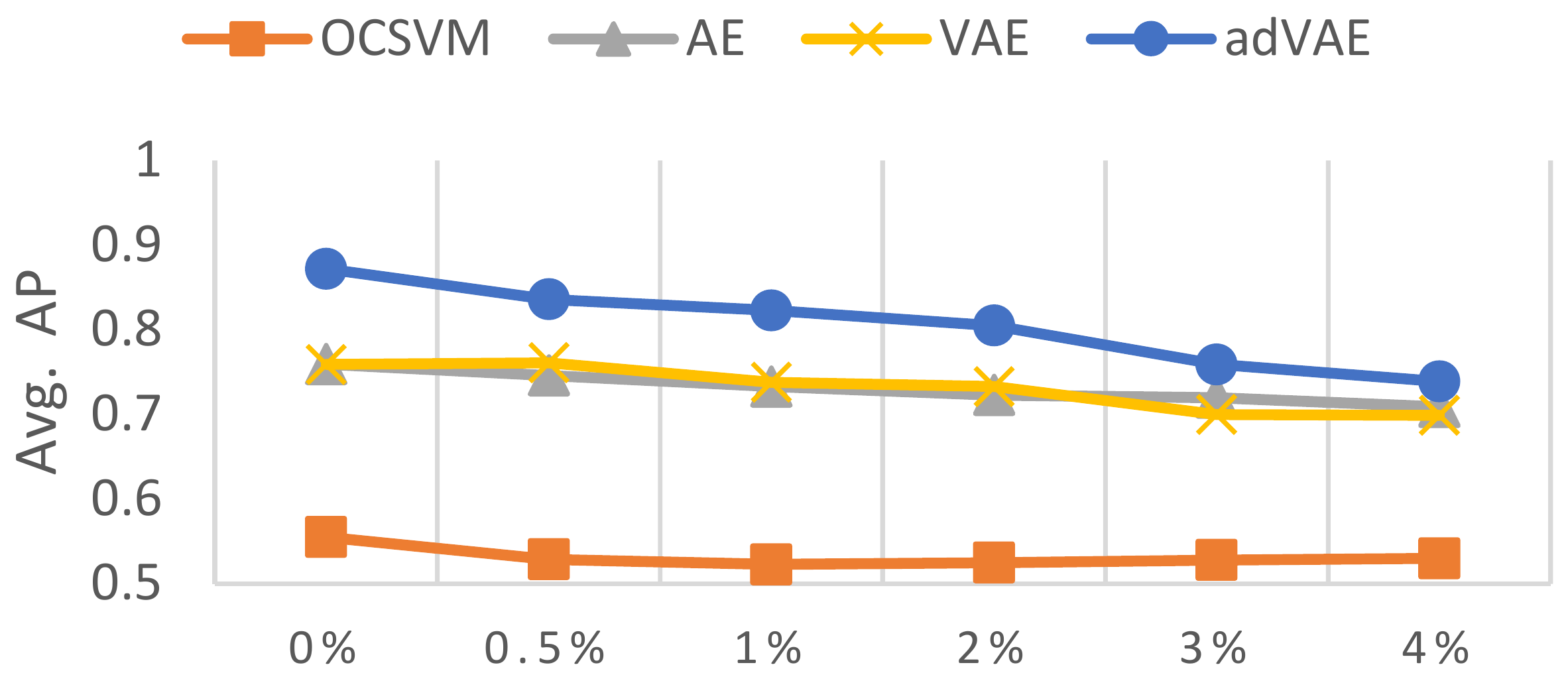}
    }
    \caption{\diff{Robustness experiment results. (a) $\lambda$ is the hyperparameter derived from VAE, which has a significant impact on performance, and hence it should e set carefully. (b) $\gamma$ and $m_z$ have a similar role: enhancing the encoder's discrimination objective. Because adVAE with $\gamma=0.01$ achieves the best results in all datasets, we suggest fixing $\gamma$ as 0.01 and tuning $m_z$ instead. (c) The model performance is not sensitive to $m_z$, and it should be considered to simply set it between 20 and 40. (d) $m_x$ is used to enhance the generator's discrimination objective, and the choice of this parameter is not very critical. For convenience, we fixed it as 2 for all tests. (e) As long as the scale of the neural network is large enough, the number of network layers or the number of neurons will not have a significant impact on the results. (f) Contaminated training data negatively affect the detection accuracy and adVAE is more sensitive, especially when the contamination ratio is larger than 3\%, because the additional discrimination objectives of adVAE are more susceptible to contamination data.}}
    \label{fig:hyperparameters}
\end{figure*}
To illustrate the impact of different designs on the model results, we design three different types of ablation experiments:

\diff{(\romannumeral1) Because adVAE is formed by adding two different training objectives to plain VAE, it is important to demonstrate how the two discrimination objectives affect the performance. Hence we propose E-adVAE and G-adVAE models (Figure~\ref{fig:E-adVAE} and \ref{fig:G-adVAE}) to demonstrate the effect of removing one of the discrimination objectives from adVAE. According to Table \ref{tab:APAUC} and \ref{tab:RecallF1}, the results show that both G-adVAE and E-adVAE achieve performance improvements compared to plain VAE, which indicates the validity of the two discriminative objectives of adVAE. In addition, E-adVAE is more competitive than G-adVAE in most datasets. This result is analyzed as follows. Because the latent variables $z$ have a lower dimension than the original data $x$, $x$ has more information than $z$. Therefore, the benefit of adding the discriminating target to the encoder $E$ is more significant than adding it to the generator $G$. }

\diff{(\romannumeral2) Because the encoder and generator work together to detect outliers, we performed a latent space visualization to prove the improvement in the encoder's anomaly detection capability.} Figure~\ref{fig:zspace} illustrates that the encoder of adVAE has a better discrimination ability, because the latent variables are separated in the latent space and more easily distinguished by the later generator. 

\diff{(\romannumeral3) To independently verify the anomaly detection capabilities of the generator, we synthesize latent outlier variables $z_{o}$ by adding five types of noise to the normal latent variables $z_{n}$. These latent variables are the outputs of the five corresponding encoders and the input data of the encoders are 200 normal samples $x_{n}$ from the pen~\cite{DBLP:conf/icde/KellerMB12} dataset. The normal scores vector $s_{n}\in \bm{R}^{200}$ can be calculated from $x_{n}$ and $G(z_{n})$, and the anomalous scores vector $s_{o}\in \bm{R}^{200}$ can be calculated from $x_{n}$ and $G(z_{o})$ (as in section \ref{score}). Afterwards, the Wasserstein distance between $s_{n}$ and $s_{o}$ is used to measure the detection capabilities of each generator. Figure~\ref{fig:G-power} indicates that the generator $G$ of adVAE also has a better ability to discriminate latent variables. }

In conclusion, benefiting from better discrimination abilities, adVAE has better anomaly detection performance. This proves that the proposed self-adversarial mechanism is a prospective way of customizing generative models to fit outlier detection tasks.
\subsection{Robustness Experiments}
\diff{Because adVAE involves four primary hyperparameters, it is essential to implement in-depth hyperparameter experiments comprehensively. As shown in Figure \ref{fig:lambda}--\ref{fig:MSE-margin}, we tested the results for four hyperparameters. Actually, only $m_z$ and $m_x$ need to be tuned, because $\lambda$ is derived from VAE and $\gamma$ is fixed as 0.01. Based on that, finding good hyperparameters of adVAE is not more complicated than plain VAE, because it is not sensitive to $m_z$ and $m_x$ according to Figure \ref{fig:KLD-margin} and \ref{fig:MSE-margin}.}

\diff{As for the learning ability of neural networks, we tune the number of layers from 2 to 6 and adjust the number of hidden neurons from $\frac{1}{3}dim(x)$ to $dim(x)$. The results are shown in Figure \ref{fig:network-structure}, where V1--V5 represent the parameter variation. It can be seen that V1 achieved a slightly worse detection capability, because the insufficient networks cannot learn enough information of the normal training data. As long as the scale of neural network is within a reasonable range, adVAE is robust to network structure.}

\diff{Because adVAE needs an anomaly-free dataset, it is important to investigate how adVAE responds to contaminated training data. Meanwhile, we also choose three semisupervised anomaly detection methods for comparison. ADASYN~\cite{DBLP:conf/ijcnn/HeBGL08} is used to synthesize new anomalous samples. The results of Figure \ref{fig:add-outliers} show that OCSVM is more robust with a contaminated dataset than the other three AE-based methods. This is because OCSVM can ignore certain noises when learning the boundary of the training data, whereas AE-based models always try to reduce the reconstruction error for all training data. Moreover, a high contamination ratio will more easily disrupt the proposed discrimination loss, which suggests training an AE-based anomaly detection model with high-quality data (i.e., a clean or low-contamination-ratio dataset). In practice, normal data is easy to collect, and thus the contamination ratio usually remains at a low level.} 
\section{Conclusion}
In this paper, we have proposed a self-\textbf{ad}versarial \textbf{V}ariational \textbf{A}uto-\textbf{e}ncoder (adVAE) with a Gaussian anomaly prior assumption and a self-adversarial mechanism. The proposed adVAE is encouraged to distinguish the normal latent code and the anomalous latent variables generated by the Gaussian transformer $T$, which can also be regarded as an outstanding regularization introduced into VAE-based outlier detection method. The results demonstrate that the proposed adVAE outperforms than other state-of-the-art anomaly detection methods. In the future, we will try to redesign the discrimination objective of the generator to further enhance the generator's ability to recognize anomalies.

\section*{Acknowledgement}
This research is in part supported by National Nature Science Foundation of China (No. 51777122).

\section*{\refname}

\bibliography{adVAE}

\begin{thebibliography}{10}
\expandafter\ifx\csname url\endcsname\relax
  \def\url#1{\texttt{#1}}\fi
\expandafter\ifx\csname urlprefix\endcsname\relax\def\urlprefix{URL }\fi
\expandafter\ifx\csname href\endcsname\relax
  \def\href#1#2{#2} \def\path#1{#1}\fi

\bibitem{osada2017network}
G.~Osada, K.~Omote, T.~Nishide, Network intrusion detection based on
  semi-supervised variational auto-encoder, in: European Symposium on Research
  in Computer Security (ESORICS), Springer, 2017, pp. 344--361 (2017).

\bibitem{DBLP:journals/jnca/AbdallahMZ16}
A.~Abdallah, M.~A. Maarof, A.~Zainal, Fraud detection system: {A} survey,
  Journal of Network and Computer Applications 68 (2016) 90--113 (2016).

\bibitem{cui2019improved}
P.~Cui, C.~Zhan, Y.~Yang, Improved nonlinear process monitoring based on
  ensemble kpca with local structure analysis, Chemical Engineering Research
  and Design 142 (2019) 355--368 (2019).

\bibitem{schlegl2017unsupervised}
T.~Schlegl, P.~Seeb{\"{o}}ck, S.~M. Waldstein, U.~Schmidt{-}Erfurth, G.~Langs,
  Unsupervised anomaly detection with generative adversarial networks to guide
  marker discovery, in: International Conference on Information Processing in
  Medical Imaging {(IPMI)}, Springer, 2017, pp. 146--157 (2017).

\bibitem{DBLP:conf/accv/AkcayAB18}
S.~Akcay, A.~A. Abarghouei, T.~P. Breckon, Ganomaly: Semi-supervised anomaly
  detection via adversarial training, in: Asian Conference on Computer Vision
  ({ACCV}), Springer, 2018, pp. 622--637 (2018).

\bibitem{DBLP:journals/kbs/ZhangBXRFQF19}
C.~Zhang, J.~Bi, S.~Xu, E.~Ramentol, G.~Fan, B.~Qiao, H.~Fujita,
  Multi-imbalance: An open-source software for multi-class imbalance learning,
  Knowl.-Based Syst. 174 (2019) 137--143 (2019).

\bibitem{zhou2019deep}
F.~Zhou, S.~Yang, H.~Fujita, D.~Chen, C.~Wen, Deep learning fault diagnosis
  method based on global optimization gan for unbalanced data, Knowl.-Based
  Syst. (2019).

\bibitem{JMLR:v18:16-365}
G.~Lema{{\^i}}tre, F.~Nogueira, C.~K. Aridas, Imbalanced-learn: A python
  toolbox to tackle the curse of imbalanced datasets in machine learning,
  Journal of Machine Learning Research 18~(17) (2017) 1--5 (2017).

\bibitem{pimentel2014review}
M.~A.~F. Pimentel, D.~A. Clifton, L.~A. Clifton, L.~Tarassenko, A review of
  novelty detection, Signal Processing 99 (2014) 215--249 (2014).

\bibitem{chalapathy2019deep}
R.~{Chalapathy}, S.~{Chawla}, {Deep Learning for Anomaly Detection: A Survey},
  arXiv e-prints (2019) arXiv:1901.03407 (2019).

\bibitem{kingma2013auto}
D.~P. {Kingma}, M.~{Welling}, {Auto-Encoding Variational Bayes}, in:
  International Conference on Learning Representations {(ICLR)}, 2014 (2014).

\bibitem{goodfellow2014generative}
I.~J. Goodfellow, J.~Pouget{-}Abadie, M.~Mirza, B.~Xu, D.~Warde{-}Farley,
  S.~Ozair, A.~C. Courville, Y.~Bengio, Generative adversarial nets, in: Annual
  Conference on Neural Information Processing Systems {(NeurIPS)}, {MIT Press},
  2014, pp. 2672--2680 (2014).

\bibitem{an2015variational}
J.~An, S.~Cho, Variational autoencoder based anomaly detection using
  reconstruction probability, Tech. rep., SNU Data Mining Center (2015).

\bibitem{8279425}
D.~{Park}, Y.~{Hoshi}, C.~C. {Kemp}, A multimodal anomaly detector for
  robot-assisted feeding using an lstm-based variational autoencoder, IEEE
  Robotics and Automation Letters 3~(3) (2018) 1544--1551 (2018).

\bibitem{suh2016echo}
S.~Suh, D.~H. Chae, H.~Kang, S.~Choi, Echo-state conditional variational
  autoencoder for anomaly detection, in: International Joint Conference on
  Neural Networks {(IJCNN)}, {IEEE}, 2016, pp. 1015--1022 (2016).

\bibitem{xu2018unsupervised}
H.~Xu, W.~Chen, N.~Zhao, Z.~Li, J.~Bu, Z.~Li, Y.~Liu, Y.~Zhao, D.~Pei, Y.~Feng,
  J.~Chen, Z.~Wang, H.~Qiao, Unsupervised anomaly detection via variational
  auto-encoder for seasonal kpis in web applications, in: International World
  Wide Web Conference {(WWW)}, {ACM}, 2018, pp. 187--196 (2018).

\bibitem{makhzani2015adversarial}
A.~Makhzani, J.~Shlens, N.~Jaitly, I.~Goodfellow, Adversarial autoencoders, in:
  International Conference on Learning Representations {(ICLR)}, 2016 (2016).

\bibitem{pidhorskyi2018generative}
S.~Pidhorskyi, R.~Almohsen, G.~Doretto, Generative probabilistic novelty
  detection with adversarial autoencoders, in: Annual Conference on Neural
  Information Processing Systems {(NeurIPS)}, {MIT Press}, 2018, pp. 6823--6834
  (2018).

\bibitem{ravanbakhsh2017abnormal}
M.~Ravanbakhsh, M.~Nabi, E.~Sangineto, L.~Marcenaro, C.~S. Regazzoni, N.~Sebe,
  Abnormal event detection in videos using generative adversarial nets, in:
  International Conference on Image Processing {(ICIP)}, {IEEE}, 2017, pp.
  1577--1581 (2017).

\bibitem{DBLP:conf/iclr/0022KSDDSSA17}
X.~Chen, D.~P. Kingma, T.~Salimans, Y.~Duan, P.~Dhariwal, J.~Schulman,
  I.~Sutskever, P.~Abbeel, Variational lossy autoencoder, in: International
  Conference on Learning Representations {(ICLR)}, 2017 (2017).

\bibitem{DBLP:conf/nips/FraccaroSPW16}
M.~Fraccaro, S.~K. S{\o}nderby, U.~Paquet, O.~Winther, Sequential neural models
  with stochastic layers, in: Annual Conference on Neural Information
  Processing Systems {(NeurIPS)}, 2016, pp. 2199--2207 (2016).

\bibitem{DBLP:conf/aaai/SerbanSLCPCB17}
I.~V. Serban, A.~Sordoni, R.~Lowe, L.~Charlin, J.~Pineau, A.~C. Courville,
  Y.~Bengio, A hierarchical latent variable encoder-decoder model for
  generating dialogues, in: AAAI Conference on Artificial Intelligence
  {(AAAI)}, 2017, pp. 3295--3301 (2017).

\bibitem{honkela2004variational}
A.~Honkela, H.~Valpola, Variational learning and bits-back coding: an
  information-theoretic view to bayesian learning, IEEE Transactions on Neural
  Networks 15~(4) (2004) 800--810 (2004).

\bibitem{8668550}
Y.~{Liu}, Z.~{Li}, C.~{Zhou}, Y.~{Jiang}, J.~{Sun}, M.~{Wang}, X.~{He},
  Generative adversarial active learning for unsupervised outlier detection,
  IEEE Transactions on Knowledge and Data Engineering (2019).

\bibitem{kawachi2018complementary}
Y.~Kawachi, Y.~Koizumi, N.~Harada, Complementary set variational autoencoder
  for supervised anomaly detection, in: International Conference on Acoustics,
  Speech and Signal Processing {(ICASSP)}, {IEEE}, 2018, pp. 2366--2370 (2018).

\bibitem{huang2018introvae}
H.~Huang, Z.~Li, R.~He, Z.~Sun, T.~Tan, Introvae: Introspective variational
  autoencoders for photographic image synthesis, in: Annual Conference on
  Neural Information Processing Systems {(NeurIPS)}, {MIT Press}, 2018, pp.
  52--63 (2018).

\bibitem{DBLP:conf/icpr/IlonenPKK06}
J.~Ilonen, P.~Paalanen, J.~Kamarainen, H.~K{\"{a}}lvi{\"{a}}inen, Gaussian
  mixture pdf in one-class classification: computing and utilizing confidence
  values, in: International Conference on Pattern Recognition {(ICPR)}, {IEEE},
  2006, pp. 577--580 (2006).

\bibitem{DBLP:conf/icpr/YeungC02}
D.~Yeung, C.~Chow, Parzen-window network intrusion detectors, in: International
  Conference on Pattern Recognition {(ICPR)}, {IEEE}, 2002, pp. 385--388
  (2002).

\bibitem{DBLP:conf/sigmod/BreunigKNS00}
M.~M. Breunig, H.~Kriegel, R.~T. Ng, J.~Sander, {LOF:} identifying
  density-based local outliers, in: ACM SIGMOD International Conference on
  Management of Data {(SIGMOD)}, {ACM}, 2000, pp. 93--104 (2000).

\bibitem{DBLP:journals/ijon/TangH17}
B.~Tang, H.~He, A local density-based approach for outlier detection,
  Neurocomputing 241 (2017) 171--180 (2017).

\bibitem{OCSVM}
B.~Sch{\"{o}}lkopf, J.~C. Platt, J.~Shawe{-}Taylor, A.~J. Smola, R.~C.
  Williamson, Estimating the support of a high-dimensional distribution, Neural
  Computation 13~(7) (2001) 1443--1471 (2001).

\bibitem{DBLP:journals/kbs/YinWF18}
L.~Yin, H.~Wang, W.~Fan, Active learning based support vector data description
  method for robust novelty detection, Knowl.-Based Syst. 153 (2018) 40--52
  (2018).

\bibitem{Olive2017}
D.~J. Olive, Principal Component Analysis, Springer, 2017, pp. 189--217 (2017).

\bibitem{DBLP:journals/candie/HarrouKCTS15}
F.~Harrou, F.~Kadri, S.~Chaabane, C.~Tahon, Y.~Sun, Improved principal
  component analysis for anomaly detection: Application to an emergency
  department, Computers {\&} Industrial Engineering 88 (2015) 63--77 (2015).

\bibitem{baklouti2016iterated}
R.~Baklouti, M.~Mansouri, M.~Nounou, H.~Nounou, A.~B. Hamida, Iterated robust
  kernel fuzzy principal component analysis and application to fault detection,
  Journal of Computational Science 15 (2016) 34--49 (2016).

\bibitem{DBLP:conf/nips/GulrajaniAADC17}
I.~Gulrajani, F.~Ahmed, M.~Arjovsky, V.~Dumoulin, A.~C. Courville, Improved
  training of wasserstein gans, in: Annual Conference on Neural Information
  Processing Systems {(NeurIPS)}, 2017, pp. 5767--5777 (2017).

\bibitem{DBLP:journals/csda/GramackiG17}
A.~Gramacki, J.~Gramacki, Fft-based fast bandwidth selector for multivariate
  kernel density estimation, Computational Statistics {\&} Data Analysis 106
  (2017) 27--45 (2017).

\bibitem{DAGMM}
B.~Zong, Q.~Song, M.~R. Min, W.~Cheng, C.~Lumezanu, D.~Cho, H.~Chen, Deep
  autoencoding gaussian mixture model for unsupervised anomaly detection, in:
  International Conference on Learning Representations {(ICLR)}, 2018 (2018).

\bibitem{silverman2018density}
B.~W. Silverman, Density estimation for statistics and data analysis,
  Routledge, 2018, Ch.~3 (2018).

\bibitem{DBLP:journals/tkdd/RayanaA16}
S.~Rayana, L.~Akoglu, Less is more: Building selective anomaly ensembles, ACM
  Trans. Knowl. Discov. Data 10~(4) (2016) 42:1--42:33 (2016).

\bibitem{DBLP:conf/sdm/SatheA16}
S.~Sathe, C.~C. Aggarwal, {LODES:} local density meets spectral outlier
  detection, in: SIAM International Conference on Data Mining {(SDM)}, {SIAM},
  2016, pp. 171--179 (2016).

\bibitem{DBLP:conf/icdm/LiuTZ08}
F.~T. Liu, K.~M. Ting, Z.~Zhou, Isolation forest, in: International Conference
  on Data Mining {(ICDM)}, {IEEE}, 2008, pp. 413--422 (2008).

\bibitem{DBLP:journals/sigkdd/AggarwalS15}
C.~C. Aggarwal, S.~Sathe, Theoretical foundations and algorithms for outlier
  ensembles, SIGKDD Explor. Newsl. 17~(1) (2015) 24--47 (2015).

\bibitem{DBLP:conf/icde/KellerMB12}
F.~Keller, E.~M{\"{u}}ller, K.~B{\"{o}}hm, Hics: High contrast subspaces for
  density-based outlier ranking, in: International Conference on Data
  Engineering {(ICDE)}, IEEE, 2012, pp. 1037--1048 (2012).

\bibitem{davis2006relationship}
J.~Davis, M.~Goadrich, The relationship between precision-recall and {ROC}
  curves, in: International Conference on Machine Learning {(ICML)}, ACM, 2006,
  pp. 233--240 (2006).

\bibitem{Iforest}
F.~T. Liu, K.~M. Ting, Z.~Zhou, Isolation forest, in: International Conference
  on Data Mining ({ICDM}), {IEEE} Computer Society, 2008, pp. 413--422 (2008).

\bibitem{ABOD}
H.~Kriegel, M.~Schubert, A.~Zimek, Angle-based outlier detection in
  high-dimensional data, in: ACM Knowledge Discovery and Data Mining ({KDD}),
  {ACM}, 2008, pp. 444--452 (2008).

\bibitem{SOD}
H.~Kriegel, P.~Kr{\"{o}}ger, E.~Schubert, A.~Zimek, Outlier detection in
  axis-parallel subspaces of high dimensional data, in: Pacific-Asia Conference
  on Knowledge Discovery and Data Mining ({PAKDD}), Vol. 5476 of Lecture Notes
  in Computer Science, Springer, 2009, pp. 831--838 (2009).

\bibitem{HBOS}
M.~Goldstein, A.~Dengel, Histogram-based outlier score (hbos): A fast
  unsupervised anomaly detection algorithm, German Conference on Artificial
  Intelligence (KI-2012): Poster and Demo Track (2012) 59--63 (2012).

\bibitem{zhao2019pyod}
Y.~Zhao, Z.~Nasrullah, Z.~Li, Pyod: A python toolbox for scalable outlier
  detection, Journal of Machine Learning Research 20~(96) (2019) 1--7 (2019).

\bibitem{DBLP:conf/ijcnn/HeBGL08}
H.~He, Y.~Bai, E.~A. Garcia, S.~Li, {ADASYN:} adaptive synthetic sampling
  approach for imbalanced learning, in: International Joint Conference on
  Neural Networks {(IJCNN)}, {IEEE}, 2008, pp. 1322--1328 (2008).

\end{thebibliography}
\bibliographystyle{elsarticle-num}


\end{document}